\documentclass[opre,nonblindrev]{informs3}

\usepackage{amssymb,amsfonts,amsmath,amscd,mathrsfs}
\usepackage{graphicx,float,psfrag,epsfig}
\usepackage{pgf,pgfarrows,pgfnodes}
\usepackage{multicol}
\usepackage{float,psfrag,epsfig}

\usepackage{wrapfig, dsfont, algorithmic, multirow, endnotes}

\OneAndAHalfSpacedXI 

\usepackage{natbib}
 \bibpunct[, ]{(}{)}{,}{a}{}{,}%
 \def\bibfont{\small}%
\TheoremsNumberedThrough     
\ECRepeatTheorems

\EquationsNumberedThrough    

\newcommand{\wparam}{{\score}}
\newcommand{\thetaparam}{\ensuremath{\theta^*}}
\newcommand{\thetastar}{\thetaparam}

\newcommand{\btilde}{\widetilde{b}}
\newcommand{\mleobs}{v}
\newcommand{\design}{x}
\newcommand{\inprod}[2]{\langle #1,#2 \rangle}
\newcommand{\thetahat}{\widehat{\theta}}
\newcommand{\Loss}{\mathcal{L}_\numobs}
\newcommand{\gradloss}{\ensuremath{\nabla \Loss}}
\newcommand{\myexp}{\inprod{\thetaparam}{\design_l}+v\inprod{\thetaparam}{\design_l}}

\renewcommand{\DS}[1]{{\color{black} {#1}}}
\newcommand{\snncomment}[1]{{\color{black} {#1}}}
\newcommand{\SN}[1]{\snncomment{#1}}

\newcommand{\ourchanges}[1]{{\color{black} {#1}}}
\newcommand{\comment}[1]{}
\newcommand{\finalcomment}[1]{{\color{black} {#1}}}

\def\cE{{\cal E}}
\def\cF{{\cal F}}

\def\cA{{\cal A}}
\def\tP{{\tilde P}}
\def\tA{{\tilde A}}
\def\tsigma{{\tilde \sigma}}
\def\tZ{{\tilde Z}}

\def\tpi{{\tilde \pi}}
\def\defn{\ensuremath{:=}}
\def\tPi{{\tilde \Pi}}

\def\tL{{\tilde L}}
\def\lmax{\lambda_{\rm max}}

\newcommand{\barDelta}{\bar{\Delta}}
\def\dmax{d_{\rm max}}
\def\dmin{d_{\rm min}}

\def\<{\langle}
\def\>{\rangle}
\def\ones{\mathds 1}
\def\prob{\mathbb P}
\def\E{\mathbb E}
\def\reals{\mathbb R}
\def\score{w}
\def\numitems{n}
\def\numobs{m}
\def\resamp{k}
\def\numcomp{d}
\def\upbound{b}
\def\ind{\mathbb I}
\def\dist{D}

\def\npack{M}
\def\idxpack{L}
\def\idxest{\hat L}
\def\distpack{\delta}
\def\KL{D_{\rm KL}}
\def\obs{X}
\def\bY{{\bar{Y}}}
\def\cS{{\cal S}}
\def\eiggap{\xi}
\def\degratio{\kappa}

\newcommand{\RC}{{\textsc Rank Centrality}}

\begin{document}



\RUNTITLE{Rank Centrality: Ranking from Pair-wise Comparisons}

\TITLE{Rank Centrality: Ranking from Pair-wise Comparisons}

\ARTICLEAUTHORS{%
\AUTHOR{Sahand Negahban}
\AFF{Statistics Department, Yale University, 24 Hillhouse Ave, New Haven, CT 06510 , \EMAIL{sahand.negahban@yale.edu} }
\AUTHOR{Sewoong Oh}
\AFF{Department of Industrial and Enterprise Systems Engineering, University of Illinois at Urbana-Champaign, 104 S. Mathews Ave., Urbana, IL 61801, \EMAIL{swoh@illinois.edu}}
\AUTHOR{Devavrat Shah\thanks{This work was supported in parts by MURI  W911NF-11-1-0036 and NSF CMMI-1462158.}}
\AFF{Department of Electrical Engineering and Computer Science, Massachusetts Institute of Technology, Massachusetts Ave., Cambridge, MA 02139, \EMAIL{devavrat@mit.edu}} 
} 

\ABSTRACT{
  The question of aggregating pair-wise comparisons to obtain a global
  ranking over a collection of objects has been of interest for a very
  long time: be it ranking of online gamers (e.g. MSR's TrueSkill
  system) and chess players, aggregating social opinions, or deciding
  which product to sell based on transactions. In most settings, in
  addition to obtaining a ranking, finding `scores' for each object
  (e.g. player's rating) is of interest for understanding the intensity 
  of the preferences.

  In this paper, we propose \RC, an iterative rank aggregation
  algorithm\footnote{\DS{Similar algorithms, based on the comparison
      data matrix have been proposed in the literature. As discussed
      in detail in Section \ref{sSec:experiment}, they are all
      different from \RC.}} for discovering scores for objects (or
  items) from pair-wise comparisons. The algorithm has a natural {\em
    random walk} interpretation over the graph of objects with an edge
  present between a pair of objects if they are compared; the score,
  which we call Rank Centrality, of an object turns out to be its
  stationary probability under this random walk.
  
  To study the efficacy of the algorithm, we consider the popular
  Bradley-Terry-Luce (BTL) model (equivalent to the Multinomial Logit
  (MNL) for pair-wise comparisons) in which each object has an
  associated score which determines the probabilistic outcomes of
  pair-wise comparisons between objects. In terms of the pair-wise
  marginal probabilities, which is the main subject of this paper, the
  MNL model and the BTL model are identical.  We bound the finite
  sample error rates between the scores assumed by the BTL model and
  those estimated by our algorithm. In particular, the number of
  samples required to learn the score well with high probability
  depends on the structure of the comparison graph.  When the
  Laplacian of the comparison graph has a strictly positive spectral
  gap, e.g.  each item is compared to a subset of randomly chosen
  items, this leads to dependence on the number of samples that 
  is nearly order-optimal.

  Experimental evaluations on synthetic datasets generated according to the BTL model 
  show that our algorithm performs as well as the Maximum Likelihood estimator for that 
  model and outperforms other popular ranking algorithms.  
  
}

\HISTORY{This paper was first submitted on December 1st, 2013.}

\maketitle


%

\section{Introduction}
Rank aggregation is an important task in a wide range of learning
and social contexts arising in recommendation systems, information
retrieval, and sports and competitions. Given $\numitems$ items, we wish to
infer relevancy scores or an ordering on the items based on partial
orderings provided through many (possibly contradictory) samples. Frequently,
the available data that is presented to us is in the form of a comparison: player $A$
defeats player $B$; book $A$ is purchased when books $A$ and $B$ are
displayed (a bigger collection of books implies multiple pair-wise comparisons);
movie $A$ is liked more compared to movie $B$. From such partial preferences in
the form of comparisons, we frequently wish to deduce not only the order of the
underlying objects, but also the scores associated with the objects themselves so 
as to deduce the intensity of the resulting preference order. 

For example, the Microsoft TrueSkill engine assigns scores to online gamers based on 
the outcomes of (pair-wise) games between players. Indeed, it assumes that
each player has inherent ``skill'' and the outcomes of the games are used to 
learn these skill parameters which in turn lead to scores associated with each player. 
In most such settings, similar model-based approaches are employed. 

In this paper, we have set out with the following goal: develop an
algorithm for the above stated problem which $(a)$ is computationally
simple, $(b)$ works with available (comparison) data only, and $(c)$ when
data is generated as per a reasonable model, then the algorithm should 
do as well as the best model aware algorithm. The main result of this paper 
is an affirmative answer to these questions.

\medskip
\noindent
{\bf Related work.} 
Most rating based systems rely on users to provide explicit
numeric scores for their interests. While these assumptions have led
to a flurry of theoretical research for item recommendations based on
matrix completion (cf.~\citet{CR09,KMO10,NW11}), \DS{arguably} 
numeric scores provided by individual users are generally inconsistent. 
\DS{Furthermore, in a number of learning contexts as illustrated above, explicit
scores are not available. }


These observations have led to the need to develop methods that can
aggregate such forms of ordering information into relevance ratings.
In general, however, designing consistent aggregation methods can be challenging due
in part to possible contradictions between individual preferences. 
For example, if we consider items $A$, $B$, and $C$, one user might prefer $A$ to
$B$, while another prefers $B$ to $C$, and a third user prefers $C$ to $A$. 
Such problems have been well studied starting with (and potentially even before) \citet{Co1785}. 
In the celebrated work by \citet{Arrow53}, existence of a rank aggregation 
algorithm with reasonable sets of properties (or axioms) was shown to be impossible. 

In this paper, we are interested in a more restrictive setting: we
have outcomes of pair-wise comparisons between pairs of items, rather
than a complete ordering as considered in
\citep{Arrow53}. 
Based on those pair-wise comparisons, we want to obtain a ranking of
items along with a score for each item indicating the intensity of the
preference. 
One reasonable way to think about our setting is to imagine
that there is a distribution over orderings or rankings or
permutations of items (also known as the {\em discrete choice model} in 
the literature on Social Choice) and every time a pair of items is compared, the
outcome is generated as per this underlying distribution.  
\ourchanges{Examples of popular distributions over permutations include 
the Plackett-Luce model \citep{Luce59,Pla75} and the Mallows model \citep{Mallows57}.}
With this,
our question becomes even harder than the setting considered by
\citet{Arrow53} as, in that work, effectively the entire
distribution over permutations was already known!

Indeed, such hurdles have not stopped the scientific community as well as practical 
designers from designing such systems. Chess rating systems and the more
recent MSR TrueSkill Ranking system are prime examples. Our work falls
precisely into this realm: design algorithms that work well in practice, makes sense 
in general, and perhaps more importantly, have attractive theoretical properties under common comparative judgment models. 
\ourchanges{An important
and landmark model in this class is called the Plackett-Luce model,
which is also known as the Multinomial Logit
(MNL) model (cf. \citet{McF73}) in the operations research and social 
science literature. A special case of the Plackett-Luce model applied to pair-wise comparisons is known as 
the Bradley-Terry-Luce (BTL) model \citep{BT55,Luce59}.}
It has been the backbone of
many practical system designs including pricing in the airline industry, e.g.
see \citet{VanRyzin}. \citet{adleretal} used such models to design adaptive algorithms that
select the winner from small number of rounds.  Interestingly enough,
the (near-)optimal performance of their adaptive algorithm for winner
selection is matched by our non-adaptive algorithm
for assigning scores to obtain global rankings of all players.

{\ourchanges{
We propose a new rank aggregation algorithm, which we call \RC, 
that builds on a long line of research in using eigenvectors of certain matrices  
to find global rankings of items, which dates back to \cite{See49}. 
This line of research is referred to as {\em spectral ranking} and 
for an extensive survey we refer to \cite{Vig09}. 
Given pair-wise comparisons of items from a single individual on all possible choices of pairs,  
\cite{Wei52} introduced a ranking algorithm based on the leading eigenvector of the matrix representing the comparisons outcome.
A slight generalization accounting for data from multiple decision makers was proposed by \cite{Ken55}. 
\cite{Kee93}, and more recent work by \citet{dwork01}, 
proposed several variations of spectral algorithms for ranking from pair-wise comparisons. 
We propose \RC~for ranking from pair-wise comparisons by   
using the leading eigenvector of a particular matrix formed by constructing a Markov chain 
corresponding to a random walk on a graph. 
Although it appears to be similar to the existing spectral ranking approaches, 
the precise form of the algorithm proposed is distinct and 
this precise form {\em does matter}: the empirical results using synthetic
data presented in Section \ref{sSec:experiment} make this clear. 
In summary, building on the classical field of spectral ranking, 
we propose a novel spectral ranking algorithm and 
provide a firm theoretical grounding by showing that 
it is a provably near-optimal estimator for a popular discrete choice model, i.e. the BTL model formally defined in Section \ref{sec:BTL}. 

Numerous spectral ranking algorithms have been proposed in the past, 
one of the most popular example being PageRank \citep{BP98}. 
However, almost invariably, the question of when one should choose to use a particular spectral ranking algorithm 
is left open. 
One notable exception is the work of \cite{AT05}, which provides a set of axioms satisfied by PageRank algorithm and prove that 
PageRank is the only rank aggregation algorithm that satisfies those particular axioms. 
Hence, it provides a  guideline for deciding when PageRank should be used,  
i.e. in applications where the specific set of axioms make sense. 
In a similar spirit, Rank Centrality is a spectral ranking algorithm with a theoretical justification 
suggesting that it should be used in applications where the BTL or MNL model makes sense 
(in the remainder of this manuscript, we shall use BTL model as representative for BTL and MNL model).  
}}

{\ourchanges{
There has been significant work on rankings from pair-wise comparison in the last several years. 
A popular model is a distribution over permutations known as the Mallows model, which assigns probability to 
observed rankings according to the 
Kendall-$\tau$ distance to a true ranking.
Since the maximum likelihood estimation is provably difficult,  
\cite{DKNS01} studied this problem (also known as the Kemeny optimization) when full rankings are observed  
and provided a 2-approximation algorithm.
This was later improved by \cite{ACN08} and also generalized to partial rankings \citep{Ail10}. 
Recently, \cite{LB11} proposed \DS{an} expectation-maximization approach with novel sampling schemes  
to learn the Mallows model from pair-wise comparisons. 
These distance-based approaches aim to provide 
good approximation algorithms for the provably difficult problem of minimizing the Kendall-$\tau$ distance and some variations of it (e.g. \cite{FTO12}). 

Learning to rank from pair-wise comparisons has also been studied in applications  
where one might observe more than just the ordinal outcome of pair-wise comparisons.    
Additional data on cardinal preferences such as the margin of victory 
(the difference between the winning team's score and the losing team's score) 
in a football match has led to 
score-based methods for ranking, where the goal is to find scores for each team such that 
the difference of the scores is consistent with the observed margins of victory \citep{Hoch06,GL11,JLYY11}. 
More recently, \cite{VZ12} proposed 
a unified model that generalizes both the BTL model and the cardinal preferences. 
These approaches add to the traditional approaches based on 
some notion of distance, such as the Kendall-$\tau$ distance, and 
probabilistic models, such as the BTL model.

Another probabilistic model directly parameterizes the distribution of pair-wise comparisons for all the pairs 
and asks the question of whether existing pair-wise ranking algorithms are consistent or not \citep{DucMacJor10, RA14}. 
It is shown that many existing algorithms do not meet the proposed `consistency' criteria and 
new regret/optimization based algorithms are presented. }}

\DS{The algorithm proposed by \citet{AS11} can be viewed as natural adaption of Borda count based on
pair-wise comparison data. They establish it to be equivalent to Borda count based on entire distribution 
when perfect pair-wise marginals are available, i.e. large sample limit. In \citet{BM08}, the
authors present an algorithm that produces an ordering based on
$O(\numitems \log \numitems)$ pair-wise comparisons on adaptively
selected pairs.  They assume that there is an underlying true ranking
and one observes noisy comparison results.  Each time a pair is
queried, we are given the true ordering of the pair with probability
$1/2 + \gamma$ for some $\gamma > 0$ which does not depend on the
items being compared.}

%
%
%

\medskip
\noindent{\bf Our contributions.} 
In this paper, we introduce \RC, 
an iterative algorithm that takes the noisy comparison answers between a
subset of all possible pairs of items as input and produces scores for
each item as the output. The proposed algorithm has a nice intuitive
explanation. Consider a graph with nodes/vertices corresponding to the
items of interest (e.g. players). Construct a random walk on this
graph where at each time, the random walk is likely to go from vertex
$i$ to vertex $j$ if items $i$ and $j$ were ever compared; and if so,
the likelihood of going from $i$ to $j$ depends on how often $i$ lost
to $j$. That is, the random walk is more likely to move to a neighbor
who has more ``wins''.  How frequently this walk visits a particular
node in the long run, or equivalently the stationary distribution, is
the score of the corresponding item.  Thus, effectively this algorithm
captures preference of the given item versus all of the others, not
just immediate neighbors: the global effect induced by transitivity of
comparisons is captured through the stationary distribution.

Such an interpretation of the stationary distribution of a Markov
chain or a random walk has been an effective measure of relative
importance of a node in wide class of graph problems, popularly known
as the {\em Network Centrality} cf. \citep{newman}.  Notable examples of
such network centralities include the random surfer model on the web
graph for the version of the PageRank \citep{BP98} which computes the
relative importance of a web page, a model of a random crawler in
a peer-to-peer file-sharing network to assign trust value to each
peer in EigenTrust \citep{KSG03} and a random walk interpretation of
Rumor Centrality that assigns likelihood to each node for being source of
information (or rumor) spread in a network graph based on the foot-print 
of infection under the Susceptible-Infected model \cite{SZ11,SZ15}.

The computation of the stationary distribution of the Markov chain
boils down to `power iteration' using transition matrix lending to a
nice iterative algorithm.  To establish rigorous properties of 
the algorithm, we analyze its performance under the BTL model 
described in Section~\ref{sec:BTL}. 

\DS{Formally, we establish the following result: given $n$ items, when
comparisons between randomly chosen $\omega(n ~\log n)$ 
pairs of items are produced as per an (unknown) underlying BTL model,
\RC ~learns the true score up to an arbitrary accuracy with high probability 
as \SN{$n\to\infty$}.
It should be noted that $\Omega(n \log n)$ is a necessary number of 
(random) comparisons for any algorithm to even produce a consistent 
ranking with high probability since with fewer edges (comparisons) the
resulting random  graph will be disconnected with positive probability.
In that sense, \RC ~is nearly order-optimal.}

In general, the comparisons may not be available between randomly
chosen pairs. Let $G = ([n], E)$ denote the graph of comparisons
between these $n$ objects with an edge $(i,j) \in E$ if and only if
objects $i$ and $j$ are compared. In this setting, we establish that
with $O(\xi^{-2}\,n\,{\sf poly}(\log n) )$ comparisons, {\ourchanges{
\RC~learns the true score of the underlying BTL model up to an
  arbitrarily small error with high probability}}. Here, $\eiggap$ is the
spectral gap for the Laplacian of $G$ and this is how the graph
structure of comparisons plays a role. Indeed, as a special case when
comparisons are chosen at random, the induced graph is Erd\"os-R\'enyi
for which $\xi$ is strictly positive, independent of $n$, \ourchanges{with high
  probability}, leading to the (order) optimal performance of the
algorithm as stated earlier.

To understand the performance of \RC~compared to the other options, 
we perform an experimental study. It shows that the performance of 
\RC~is identical to the ML estimation of the BTL model. 
\DS{Furthermore, it  outperforms other popular choices.}
In summary, \RC~$(a)$ is computationally simple, $(b)$ always produces a solution 
using available data, and $(c)$ has near optimal performance with respect to a 
reasonable generative model.


Some remarks about our analytic technique. Our analysis boils down to
studying the induced stationary distribution of the random walk or
Markov chain corresponding to the algorithm. Like most such scenarios,
the only hope to obtain meaningful results for such `random noisy'
Markov chain is to relate it to stationary distribution of a {\em
  known} Markov chain. Through recent concentration of measure results for
random matrices and comparison technique using Dirichlet forms for
characterizing the spectrum of reversible/self-adjoint operators,
along with the known expansion property of the random graph, we obtain
the eventual result. Indeed, it is the consequence of such \DS{existing} powerful
results that lead to near-optimal analytic results for random comparison 
model and characterization of the algorithm's performance for general 
setting. 

As an important comparison, we provide analysis of sample complexity required 
by the maximum likelihood estimator (MLE) using the state-of-art analytic techniques, 
cf. \citet{NW11}. \finalcomment{Subsequent to our work, \cite{HOX14} extended 
our analysis of MLE and established that MLE also achieves near-optimal performance 
guarantees (up to a logarithmic factor) as well. Our numerical experiments suggests 
something even stronger,  the resulting error is effectively identical for both  MLE and \RC.}

\noindent{\bf Organization.} 
The remainder of the paper is organized as follows. In Section \ref{SecBack}, we describe the model, problem statement and the rank Centrality algorithm. Section \ref{SecMain} describes the main results -- the key theoretical properties of rank Centrality as well as it's empirical performance in the context of two real datasets from NASCAR and One Day International (ODI) cricket. 
We provide comparison of the Rank Centrality with the maximum likelihood estimator using the existing analytic techniques in the same section. We derive the Cramer-Rao lower bound on the square error for estimating parameters by any algorithm - across range of parameters, the performance of Rank Centrality and MLE matches the lower bound implied by Cramer-Rao bound as explained in Section \ref{SecMain} as well. Finally, Section \ref{SecProofs} details proofs of all 
results. We discuss and conclude in Section \ref{SecDiscuss}.

\noindent{\bf Notation.} 
In the remainder of this paper, we use $C$,
$C'$, etc. to denote absolute constants, and their value
might change from line to line.  We use $A^T$ to denote the transpose
of a matrix.  The Euclidean norm of a vector is denoted by
$\|x\|=\sqrt{\sum_ix_i^2}$, and the operator norm of a linear operator
is denoted by $\|A\|_2=\max_x x^TAx/x^Tx$.   \finalcomment{When we say {\em with high
  probability}, we mean that the probability of a sequence of events
$\{\cA_n \}_{ n=1}^\infty$ goes to one as $n$ grows: $\lim_{n\to \infty}\prob(\cA_n)=1$.}
Also define $[n] = \{1,2,\hdots,n\}$ to be the set of all integers
from $1$ to $n$.

\section{Model, Problem Statement and Algorithm}
\label{SecBack}



\subsection{Model} 
\label{sec:BTL}

In this section, we discuss a model of comparisons between various
items. This model will be used to analyze the Rank Centrality algorithm. 

%

\medskip 
\noindent {\bf Bradley-Terry-Luce model for comparative
  judgment.}  When comparing pairs of items from $n$ items of
interest, represented as $[n] = \{1,\dots, n\}$, the
Bradley-Terry-Luce model assumes that there is a weight or score
$\score_i \in \reals_+ \equiv \{x \in \reals: x > 0\}$ associated with each item $i \in [\numitems]$.  The outcome of a
comparison for pair of items $i$ and $j$ is determined only by the
corresponding weights $\score_i$ and $\score_j$.  Let $Y^l_{ij}$
denote the outcome of the $l$-th comparison of the pair $i$ and $j$,
such that $Y^l_{ij}=1$ if $j$ is preferred over $i$ and $0$ otherwise.
Then, according to the BTL model,
\begin{eqnarray*}
	Y^l_{ij} = \left\{
	\begin{array}{rl} 
	    1 &\text{ with probability } \frac{\score_j}{\score_i+\score_j}\;,\\
	    0 &\text{ otherwise}\;.
	\end{array}
    \right.
\end{eqnarray*}
Furthermore, conditioned on the score vector \finalcomment{$\score=(\score_1,\ldots, \score_n)^T$},  
it is assumed that the random variables $Y^l_{ij}$'s are independent of one another 
for all $i$, $j$, and $l$. 

Since the BTL model is invariant under the scaling of the scores, an
$n$-dimensional representation of the scores is not unique. 
Indeed, under the BTL model, a score vector $w \in \reals_{+}^{\numitems}$ 
is the equivalence class $[w]=\{w'\in\reals_{+}^{\numitems}|w'=a\,w,\text{ for some $a>0$}\}$. 
The outcome of a comparison only depends on the equivalence class of 
the score vector. 

To get a unique representation, we represent each equivalence class by
its projection onto the standard orthogonal simplex such that
$\sum_iw_i=1$.  This representation naturally defines a distance between two
equivalent classes as the Euclidean distance between two projections:
\begin{eqnarray*}
	d(w,w') &\equiv& \Big\|\frac{1}{\<w,\ones\>}w-\frac{1}{\<w',\ones\>}w'\Big\|\;.
\end{eqnarray*}
Our main result provides an upper bound on the 
(normalized) distance between the estimated score vector 
and the true underlying score vector. 

\medskip \noindent{\em \finalcomment{Bradley-Terry-Luce is equal to pair-wise marginals of Multinomial Logit (MNL)/Plackett-Luce.}}
We take a brief 
detour to remind the reader that the BTL model is identical to the MNL model in the sense that
the pair-wise distributions between objects induced under BTL are identical to that under MNL. 
Consider an equivalent way to describe an MNL model. Each object $i$ has an associated 
score $w_i > 0$. A random ordering over all $n$ objects is drawn as follows: iteratively 
fill the ordered positions $1,\dots, n$ by choosing object $i(k)$ for position $k$, amongst the
remaining objects (not chosen in the first $1, \dots, k-1$ positions) with probability proportional
to it's weight $w_{i(k)}$. It can be easily verified that in the random ordering of $n$ objects
generated as per this process, $i$ is ranked higher than $j$ with probability $w_i/(w_i + w_j)$.

\medskip \noindent {\bf Sampling model.}  We also assume that we perform a
fixed $k$ number of comparisons for all pairs $i$ and $j$ that are
considered (e.g. a best of $k$ series). This assumption is mainly to
simplify notations, and the analysis as well as the algorithm easily
generalizes to the case when we might have a different number of
comparisons for different pairs.  Given observations of pair-wise
comparisons among $\numitems$ items according to this sampling model,
we define a {\em comparisons graph} $G=([\numitems],E,A)$ as a graph
of $\numitems$ items where two items are connected if we have
comparisons data on that pair and $A$ denotes the weights on each of
the edges in $E$.

\subsection{Rank Centrality}
\label{Sec:Approach}
In our setting, we will assume that $a_{ij}$ represents the fraction
of times object $j$ has been preferred to object $i$, for example the
fraction of times chess player $j$ has defeated player $i$. Given the
notation above, we have that $a_{ij} = ({1}/{\resamp})
\sum_{l=1}^\resamp Y_{ij}^l$. Consider a random walk on a weighted
directed graph $G=([n],E,A)$, where a pair $(i,j)\in E$ if and only if
the pair has been compared. The weight edges are defined based on the
outcome of the comparisons: $A_{ij}=a_{ij}/(a_{ij}+a_{ji})$ and
$A_{ji}=a_{ji}/(a_{ij}+a_{ji})$ (note that $a_{ij} +
  a_{ji}=1$ in our setting).  We let $A_{ij}=0$ if the pair has not
been compared. Note that by the Strong Law of Large Numbers, as the
number $\resamp \to \infty$ the quantity $A_{ij}$ converges to
$\score_j/(\score_i + \score_j)$ almost surely.

A random walk can be represented by a time-independent transition
matrix $P$, where $P_{ij}=\prob(X_{t+1}=j|X_t=i)$.  By definition, the
entries of a transition matrix are non-negative and satisfy
$\sum_jP_{ij}=1$.  One way to define a valid transition matrix of a
random walk on $G$ is to scale all the edge weights by $1/\dmax$,
where we define $\dmax$ as the maximum out-degree of a node.  This
rescaling ensures that each row-sum is at most one.  Finally, to
ensure that each row-sum is exactly one, we add a self-loop to each
node.  Concretely,
\begin{eqnarray}
    \label{eq:defP}
    P_{ij}&=&\left\{
        \begin{array}{rl} 
	    \frac1\dmax A_{ij}&\text{ if }i\neq j\;,\\
	    1-\frac{1}{\dmax}\sum_{k\neq i} A_{ik}&\text{ if }i=j\;.
	\end{array}
    \right.
\end{eqnarray}
The choice to construct our random walk as above is not arbitrary.  In
an ideal setting with infinite samples ($\resamp \to \infty$) per
comparison the transition matrix $P$ would define a reversible Markov
chain under the BTL model. Recall that a Markov chain is reversible if
it satisfies the \emph{detailed balance equation}: there exists $v \in
\reals_+^\numitems$ such that $v_i P_{ij} = v_j P_{ji}$ for all $i,j$;
and in that case, $\pi \in \reals_+^\numitems$ defined as $\pi_i =
v_i/(\sum_j v_j)$ is its unique stationary distribution.  In the
ideal setting (say $k\to \infty$), we will have $P_{ij} = \tP_{ij}
\equiv (1/\dmax) \score_j/(\score_i+\score_j)$. That is, the random
walk will move from state $i$ to state $j$ with probability
proportional to the chance that item $j$ is preferred to item $i$. In
such a setting, it is clear that $v = \score$ satisfies the
reversibility conditions. Therefore, under these ideal conditions it
immediately follows that the vector $\score/\sum_i \score_i$ acts as a
valid stationary distribution for the Markov chain defined by $\tP$,
the ideal matrix. Hence, as long as the graph $G$ is connected and at
least one node has a self loop then we are guaranteed that our graph
has a unique stationary distribution proportional to $\score$. If the
Markov chain is reversible then we may apply the spectral analysis of
self-adjoint operators, which is crucial in the analysis of the
behavior of the method.

In our setting, the matrix $P$ is a noisy version (due to finite sample error) 
of the ideal matrix $\tP$ discussed above. Therefore, it naturally suggests the 
following algorithm as a surrogate.  We estimate the probability distribution 
obtained by applying matrix $P$ repeated starting from any initial condition. 
Precisely, let $p_t(i) = \prob(X_t=i)$ denote the distribution of the 
random walk at time $t$ with $p_0 = (p_0(i)) \in \reals_+^n$ be an arbitrary 
starting distribution on $[n]$. Then, 
\begin{eqnarray}
   p_{t+1}^T&=&p_t^TP\;.\label{eq:MarkovChain}
\end{eqnarray}
\DS{When the transition matrix has a unique left largest eigenvector, then starting from any initial distribution 
$p_0$, the limiting distribution $\pi$ is unique.}
This stationary distribution $\pi$ is the top left eigenvector of $P$, which makes computing $\pi$ a simple 
eigenvector computation.  
Formally, we state the algorithm, which assigns numerical scores to each node, which we
shall call \RC:
\vspace{0.2cm}
\begin{center}
\begin{tabular}{ll}
\hline
\vspace{-.35cm}\\
\multicolumn{2}{l}{\RC}\\
\hline
\vspace{-.35cm}\\
\multicolumn{2}{l}{{\bf Input:} $G=([n],E,A)$} \\
\multicolumn{2}{l}{{\bf Output:} rank $\{\pi(i)\}_{i\in[n]}$}\\
1:  & Compute the transition matrix $P$ according to \eqref{eq:defP};\\
2:  & Compute the stationary distribution $\pi$ (as the limit of \eqref{eq:MarkovChain}).\\
\hline
\end{tabular}
\end{center}
\vspace{0.2cm}

The stationary distribution of the random walk is a fixed point of the following equation: 
\begin{eqnarray*}
    \pi(i) = \sum_{j}\pi(j)\frac{A_{ji}}{\sum_\ell A_{i\ell}}\;.
\end{eqnarray*}
This suggests an alternative intuitive justification: an object receives a high rank 
if it has been preferred to other high ranking objects or if it has been preferred to 
many objects.

One key question remains: does $P$ have a well defined unique stationary
distribution? Since the Markov chain has a finite state space, there is 
always a stationary distribution or solution of the above stated fixed-point 
equations. However, it may not be unique if the Markov chain $P$ is not 
irreducible. The irreducibility follows easily when the graph is connected 
and for all edges $(i,j) \in E$, $a_{ij} > 0$, $a_{ji} > 0$. Interestingly enough,
we show that the iterative algorithm produces a meaningful solution with near 
optimal sample complexity as stated in Theorem~\ref{thm:mainER} when 
the pairs of objects that are compared are chosen at random.


\section{Main Results}
\label{SecMain}

The main result of this paper derives sufficient conditions under which the
proposed iterative algorithm finds a solution that is close to the true solution (under
the BTL model) for \DS{general model with arbitrary connected comparison graph $G$.} This 
result is stated as Theorem \ref{thm:maingeneral} below. In words, the 
result implies that to learn the true score correctly as per our algorithm, it is
sufficient to have number of comparisons scaling as $O(\xi^{-2}\,n\,{\sf poly}(\log n) )$
where $\xi$ is the spectral gap of the Laplacian of the graph $G$. This result explicitly
identifies the role played by the graph structure in the ability of the algorithm to 
learn the true scores. 

In the special case, when the pairs of objects to be compared are
chosen at random, that is the induced $G$ is an Erd\"os-R\'enyi random
graph, the \snncomment{spectral gap $\xi$ can be lower-bounded by a
  constant with high probability} and hence the resulting number of
comparisons required scales as $O(n {\sf poly}(\log n))$. This is
effectively the optimal sample complexity.


The bounds are presented as the rescaled Euclidean
norm between our estimate $\pi$ and the underlying stationary
distribution \finalcomment{of} $\tP$. This error metric provides us with a means to
quantify the relative certainty in guessing if one item is preferred
over another. 

After presenting our main theoretical result, we describe illustrative 
simulation results. We also present application of the algorithm in 
the context of two real data-sets: results of NASCAR race for ranking
drivers, and results of One Day International (ODI) Cricket for ranking
teams. 
\finalcomment{We shall discuss relation between Rank Centrality, the
maximum likelihood estimator and the information theoretic lower bound to conclude that 
both MLE and Rank Centrality are near-optimal when the pairs are chosen according to 
the Erd\"os-Renyi random graph. }


\subsection{Rank Centrality: Error bound for general graphs}

Recall that in the general setting, each pair of objects or items are
chosen for comparisons as per the comparisons graph $G([n],E)$. For
each such pair, we have $k$ comparisons available. The result below
characterizes the performance of Rank Centrality for such a general
setting.

Before we state the result, we present a few necessary notations. Let
$d_i$ denote the degree of node $i$ in $G$; let the max-degree be
denoted by $\dmax \equiv \max_i d_i$ and min-degree be denoted by
$\dmin \equiv \min_i d_i$; let $\degratio\equiv \dmax/\dmin$.  
{\ourchanges{
The {\em random walk normalized Laplacian matrix} of the graph $G$ is defined as $L=D^{-1}B$ where $D$
is the diagonal matrix with $D_{ii}=d_i$ and $B$ is the adjacency
matrix with $B_{ij} = B_{ji} = 1$ if $(i,j) \in E$ and $0$
otherwise. This normalized Laplacian, defined thus, can be thought of as a
transition matrix of a reversible random walk on graph $G$: from each
node $i$, jump to one of its neighbors $j$ with equal
probability. Given this, it is well known that the random walk normalized Laplacian of the
graph has real eigenvalues denoted as
\begin{align}
-1 &\; \leq \;\lambda_{n}(L) \;\leq \;\dots\; \leq\; \lambda_1(L) \;=\; 1. 
\end{align}
We shall denote the {\em spectral gap} of the Laplacian as 
$$\eiggap \;\;\equiv\;\; 1-\lmax(L)\;,$$
where 
\begin{align}
  \lmax(L)  &\;\;\equiv\;\; \max\{\lambda_2(L),-\lambda_n(L)\}\;.\label{eq:defLmax}
\end{align}
There is one-to-one correspondence between the eigenvalues of the random walk normalized Laplacian $L$ 
and the standard (symmetric) normalized Laplacian ${\mathbb I}-D^{-1/2}BD^{-1/2}$. 
}}
Now we state the result establishing the performance of Rank Centrality. 
\begin{theorem}
  \label{thm:maingeneral}
  Given $n$ objects and a \DS{connected} comparison graph $G=([n],
  E)$, let each pair $(i,j) \in E$ be compared for $k$ times with
  outcomes produced as per a BTL model with parameters $w_1,\dots,
  w_n$.  Then, for some positive constant $C \geq 8$ and when
  $k \geq 4 C^2 (b^5 \degratio^2/\dmax\eiggap^2)\log n$, 
  the following bound on the normalized error holds with probability
  at least $1-4n^{-C/8}$:
  \begin{eqnarray*}
    \frac{\big\|\pi-\tpi\big\|}{\|\tpi\|}&\leq& \frac{Cb^{5/2}\degratio}{\eiggap} \sqrt{\frac{\log n}{k\,\dmax}}\;,
  \end{eqnarray*}
  where $\tpi(i)=w_i/\sum_\ell w_\ell$, $b\equiv
  \max_{i,j}w_i/w_j$, and $\degratio\equiv \dmax/\dmin$.
  \end{theorem}

\subsection{Rank Centrality: Error bound for random graphs}

Now we consider the special case when the comparison graph $G$ is an
Erd\"os-R\'enyi random graph with pair $(i,j)$ being compared with
probability $\numcomp/\numitems$. When $\numcomp$ is poly-logarithmic
in $\numitems$, we provide a strong performance
guarantee. Specifically, the result stated below suggests that with
$O(n {\sf poly}(\log n))$ comparisons, Rank Centrality manages to
learn the true scores with high probability.
\begin{theorem}
  \label{thm:mainER}
  Given $n$ objects, let the comparison graph $G=([n], E)$ be
  generated by selecting each pair $(i,j)$ to be in $E$ with
  probability $d/n$ independently of everything else.  Each such
  chosen pair of objects is compared $k$ times with the outcomes of
  comparisons produced as per a BTL model with parameters $w_1,\dots,
  w_n$.  Then, if  $d\geq 10 C^2 \log n$ and $k\,d \geq 128 C^2 b^5 \log
  n$, the following bound on the error rate holds with probability at
  least $1-10 n^{-C/8}$:
  \begin{eqnarray*}
    \frac{\big\|\pi-\tpi\big\|}{\|\tpi\|}&\leq& 8 C b^{5/2} \sqrt{\frac{\log n}{k\,d}}\;,
  \end{eqnarray*}
  where $\tpi(i)=w_i/\sum_\ell w_\ell$ and $b\equiv \max_{i,j}w_i/w_j$.
\end{theorem}

\medskip
\noindent{\bf Remarks.} 
Some remarks are in order. First, Theorem \ref{thm:mainER}
\SN{immediately implies that as long as $k d$ grows super-linear in
  $\log n$, then the error goes to $0$. Furthermore, in the context
  that the number of items $n$ goes to $\infty$} as long as we choose
  \DS{$d = \Omega(\log n)$ and $kd = \omega(\log n)$, the relative
    error goes to $0$ as $n\to\infty$ with high probability. That is,
    with $\omega(n \log n)$ total samples, the relative error goes to
    $0$ with high probability. It is well-known that for Erd\"os-Renyi
    graphs, the induced graph $G$ is connected with high probability
    only when $d = \Omega(\log n)$, i.e.  when total number of pairs
    sampled scales as $\Omega(n \log n)$. Thus, \RC~is nearly
    order-optimal in this setting.}

Second, the $b$ parameter should be treated as constant. It is the
{\em dynamic} range in which we are trying to resolve the uncertainty
between scores. \SN{We are considering a regime that there
exists some uncertainty in the samples. Otherwise, if the weight
of a single item where an order $n$ greater than the weights of
other items, then it would effectively be preferred with certainty. Hence,
we would remove it from the items under consideration.}

Third, for a general graph, Theorem \ref{thm:maingeneral} implies that
by choice of $k \dmax = O(\kappa^2 \xi^{-2} \log
n)$, 
Rank Centrality learns a score vector close to the true scores with
high probability.  That is, effectively the Rank Centrality algorithm
requires $O(n \kappa^2 \xi^{-2} {\sf poly}(\log n))$ comparisons to
learn scores well. Ignoring $\kappa$, the graph structure plays a role
through $\xi^{-2}$, the squared inverse of the spectral gap of
Laplacian of $G$, in dictating the performance of Rank Centrality.  A
reversible natural random walk on $G$, whose transition matrix is the
Laplacian, has its mixing time scaling as $\xi^{-2}$ (precisely,
relaxation time). In that sense, the mixing time of natural random
walk on $G$ ends up playing an important role in the ability of Rank
Centrality to learn the true scores. \ourchanges{Hence, if one has the option
  to choose which pairs to compare, our analysis in Theorem
  \ref{thm:maingeneral} suggests that one should choose pairs such
  that the resulting graph has large spectral gap.  Spectral gap of
  the comparisons graph also plays an important role in \cite{OBO13},
  where the goal is to choose pairs to compare under a different model
  where cardinal preferences (as opposed to ordinal preferences) are
  observed.}

\SN{Finally, if we wish to obtain a relative accuracy of $\epsilon$
  with probability at least $1-\delta$ for a fixed number of items
  $n$, then our results also show that we require $k \, d \geq 512 \, b^5/\epsilon^2 \max(\log^2 (10/\delta) / \log n, \log n)$.}

\subsection{Experimental Results}\label{sSec:experiment}

Under the BTL model, define an error metric of an estimated ordering $\sigma$ 
as the weighted sum of pairs $(i,j)$ whose ordering is incorrect: 
\begin{eqnarray*}
	\dist_w(\sigma) &=& \Big\{\frac{1}{2\numitems\|w\|^2}\sum_{i<j} (w_i-w_j)^2 \,\ind\big((w_i-w_j)(\sigma_i-\sigma_j)>0\big)\Big\}^{1/2}\;,
\end{eqnarray*}
where $\ind(\cdot)$ is an indicator function. 
This is a more natural error metric compared to the Kemeny distance, 
which is an unweighted version of the above sum, 
since $\dist_w(\cdot)$ is less sensitive to errors between pairs with similar weights. 
Further, assuming without loss of generality that $w$ is normalized such that $\sum_i w_i=1$, 
the next lemma connects the error in $\dist_w(\cdot)$ to the bound provided in Theorem \ref{thm:mainER}. 
Hence, the same upper bound holds for $\dist_w$ error. 
A proof of this lemma is provided in the Appendix. 
\begin{lemma}
	\label{lem:metric}
	Let $\sigma$ be an ordering of $\numitems$ items induced by a scoring $\pi$. Then, 
	\[\dist_w(\sigma) \;\leq\; \frac{\|w-\pi\|}{\|w\|}\;.\]
\end{lemma}

\begin{figure}[h]
  \begin{center}
	\includegraphics[width=.47\textwidth]{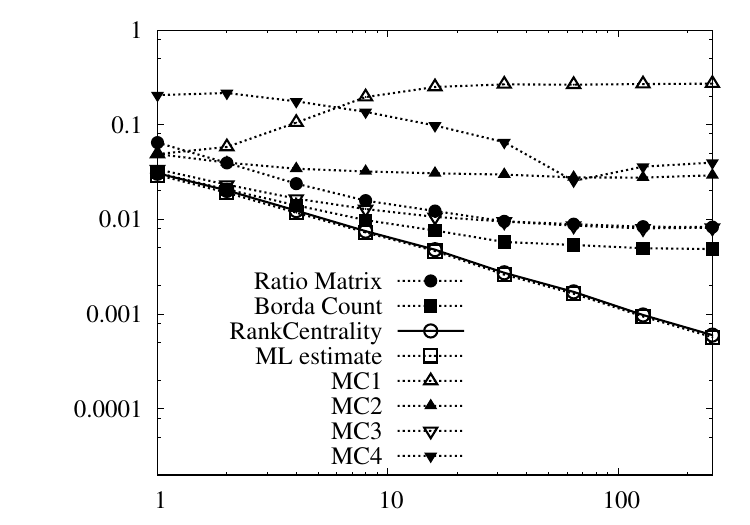}
	\includegraphics[width=.47\textwidth]{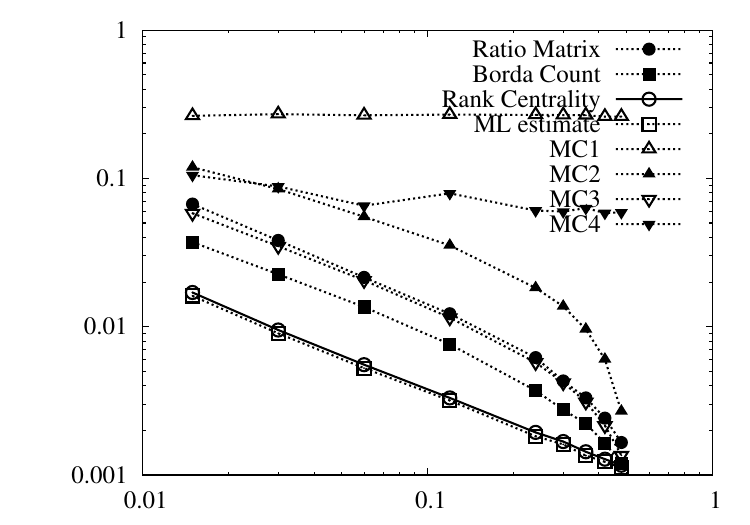}
	\put(-448,75){$\dist_w(\sigma)$}
	\put(-312,-5){$\resamp$}
	\put(-98,-5){$\numcomp/\numitems$}
  \end{center}
  \caption{\small{Average error $\dist_w(\sigma)$ of various rank aggregation algorithms averaged over $20$ instances. 
  In the figure on the left, $\numcomp$ and $\numitems$ are fixed while $\resamp$ is increased. The figure on the right, keeps
  $\resamp = 32$ fixed, and lets $d$ increase.}}
  \label{fig:experiment}
\end{figure}

\medskip
\noindent{\bf Synthetic data.} To begin with, we generate data synthetically as per a BTL model for
a specific choices of scores. 
\ourchanges{For a given $n$ and $b$, the scores are chosen such that the ratio between two consecutive scores are fixed to be $b^{1/n}$, i.e. $w_1= b^{(1-n)/2n}$, $w_2=b^{(3-n)/2n}$, $w_3=b^{(5-n)/2n}$ etc. }
A representative result is depicted in Figure.~\ref{fig:experiment}: for fixed 
$\numitems=400$ and a fixed $\upbound=10$, it shows how the error scales when varying two 
key parameters -- varying the number of comparisons per pair with fixed $d=10\log\numitems$ (on left), 
and varying the sampling probability with fixed $k=32$ (on right). This figure compares performance of
Rank Centrality with variety of other algorithms. Next, we provide a brief description of various algorithms that
we shall compare with. 

\medskip
\noindent{\em Regularized Rank Centrality.}  When there are items that have been compared only a few times, 
the scores to those items might be sensitive to the randomness in the outcome of the comparisons, 
or even worse the resulting comparisons graph might not be connected. 
To make the random walk irreducible and get a ranking that is more robust against 
comparisons noise in those edges with only a few comparisons, one can add regularization to Rank Centrality. 
A reasonable way to add regularization is to consider the transition probability $P_{ij}$ as 
the prediction of the event that $j$ beats $i$, given data $(a_{ij}, a_{ji})$. The Rank Centrality, 
in non regularized setting, uses the Haldane prior of Beta$(0,0)$, which gives 
$P_{ij} \propto a_{ij}/(a_{ij}+a_{ji})$. To add regularization, one can use different priors, 
for example Beta$(\varepsilon,\varepsilon)$, which gives 
 \begin{eqnarray}
	P_{ij} & = & \frac{1}{\dmax}\frac{a_{ij}+\varepsilon}{a_{ij}+a_{ji}+2\varepsilon}\;.
	\label{eq:regularized}
\end{eqnarray}
{When the prior is unknown, a reasonable choice in practice is $\varepsilon=1$.}

\medskip
\noindent{\em Maximum Likelihood Estimator (MLE).} The ML estimator directly maximizes the likelihood assuming the BTL model \citep{Ford57}. If we reparameterize the
problem so that $\theta_i = \log(\score_i)$ then we obtain our estimates $\widehat{\theta}$
by solving the convex program
\begin{equation}
  \widehat{\theta} \in \arg \min_{\theta} \sum_{(i,j) \in E} \sum_{l=1}^\resamp \log(1+\exp(\theta_j - \theta_i)) - Y_{ij}^l (\theta_j - \theta_i),
  \label{eq:logitregression}
\end{equation}
which is pair-wise logistic regression. The MLE is known to be consistent \citep{Ford57}. The finite sample analysis 
of MLE is provided in Section \ref{sec:ML}. 

\medskip
\noindent For comparison with Regularized Rank Centrality, we provide regularized MLE or regularized Logistic Regression:
\begin{eqnarray}
	 \arg \min_{\theta} \;\;\;\;\; \sum_{(i,j)\in E} \sum_{l} \Big\{\log(1+\exp(\theta_{j}-\theta_{i}))-Y^l_{ij}(\theta_j-\theta_i)\Big\} + \frac12 \lambda \|\theta\|^2 
	\label{eq:logit}
\end{eqnarray}

\medskip
\noindent{\em Borda Count.} \ourchanges{The (generalized) Borda Count method, analyzed recently by \citet{AS11}, scores an item 
by counting the number of wins  divided by the total number of comparisons: 
\begin{eqnarray*}
	s(i) &=& \frac{\# \text{ of times item $i$ has won} }{\# \text{ of times item $i$ has been compared }}\;. 
\end{eqnarray*}
This can be thought of as 
an extension of the standard Borda Count for  aggregating full rankings \citep{Borda1781}, which is widely used in psychology 
\citep{Dav63,KS40,Mos51}.   
If we break the full rankings into 
pair-wise comparisons and apply the pair-wise version of the Borda Count from  \citep{AS11},  then it produces 
the same ranking as the standard  Borda Count applied to the original full rankings. 
This is different from how 
HodgeRank from \cite{JLYY11} generalizes Borda count, which  does not normalize the scores by the number of comparisons. }

\medskip
\noindent {\em Spectral Ranking Algorithms.}  Rank Centrality can be classified as part of the spectral ranking 
algorithms, which assign scores to the items according to the leading eigenvector of a matrix that 
represents the data. Different choices of the matrix based on data can lead to different algorithms. 
Few prominent examples are {\em Ratio matrix} in \citep{Saa03} and those in \cite{dwork01}. 
In Ratio matrix algorithm, 
a matrix $M\in\reals^{n\times n}$ with $M_{ij}=a_{ij}/a_{ji}$ is constructed (and $M_{ii}=1$), and 
the scores for the times are assigned as per the top eigenvector of this ratio matrix.  
\cite{dwork01} introduced four spectral ranking algorithms called
MC1, MC2, MC3 and MC4. 
They are all based on a random walk very similar (but distinct) 
to that of Rank Centrality. \ourchanges{
These algorithms use the stationary distributions of the following Markov chains 
respectively, 
translated to account for the pair-wise comparisons data:  
$P^{\rm (MC1)}_{ij} = 1/|\{\ell : a_{i\ell }>0 \}|$, $P_{ij}^{\rm (MC2)} = a_{ij}/\sum_{\ell\neq i} a_{i\ell}$,  
\begin{eqnarray*}
	P^{\rm (MC3)}_{ij} = \left\{ 
		\begin{array}{rl}
		 a_{ij}/{\rm deg(i)} & \text{ if } i\neq j \;. \\ 
		 1-\sum_{\ell \neq i} a_{i \ell}/{\rm deg(i)} &\text{ if } i=j\;.
		\end{array}
		\right.
	\;,\;\; 
	 P_{ij}^{\rm (MC4)} = \left\{ 
		\begin{array}{rl}
		 1/n & \text{ if } a_{ij} \geq a_{ji}  \;,\\
		 0 & \text{ if } a_{ij} < a_{ji}  \;, \\
		 1-\sum_{\ell \neq i} |\{\ell:a_{i\ell} \geq a_{\ell i}\}| /n & \text{ if } i=j \;,
		\end{array}
		\right.
\end{eqnarray*}
where deg($i)$ is the number of items that item $i$ has been compared to. }

We make note of the following observations from Figure \ref{fig:experiment}. First, the error achieved by our Rank Centrality 
is comparable to that of ML estimator, and vanishes at the rate of $1/\sqrt{\resamp}$ as predicted by our main result. Moreover,
as predicted by our bounds, the error scales as $1/\sqrt{\numcomp}$.  Second,  for fixed $\numcomp$, both the Borda Count 
and Ratio Matrix algorithms have strictly positive error even if we take $\resamp\to\infty$. This exhibits that these are 
inherently inefficient algorithms. Third, despite strong similarity between Rank Centrality and the Markov chain based algorithms 
of \cite{dwork01}, the careful choice of the transition matrix of Rank Centrality makes a noticeable difference as shown in the 
figure - like Borda count and Ratio matrix, for fixed $d, n$, despite $k$ increasing the error remains finite (and at times gets worse!).

\medskip
\noindent{\bf Real data-sets.} \finalcomment{Next we show that Rank Centrality is more robust to 
randomly missing data compared to 
existing spectral ranking approaches on real datasets, 
which are not necessarily derived from the BTL model. } 

\medskip
\noindent {\em Dataset 1: Washington Post.} This is the public dataset collected from an online polling on 
Washington Post{\footnote {\tt http://www.washingtonpost.com/wp-srv/interactivity/worst-year-voting.html}} 
from December 2010 to January 2011.  Using allourideas{\footnote {\tt http://www.allourideas.org}} platform 
developed by \citet{SL12}, they asked who had the worst year in Washington, where each user was asked 
to compare a series of randomly selected pairs of political entities. There are $67$ political entities in the 
dataset, and the resulting graph is a complete graph on these $67$ nodes. We used Rank Centrality and 
other algorithms to aggregate this data. We use this data-set primarily to check the 'robustness' of algorithms
rather than understanding their ability to identify ground truth as by design it is not available. 

\begin{figure}[h]
  \begin{center}
	\includegraphics[width=.7\textwidth]{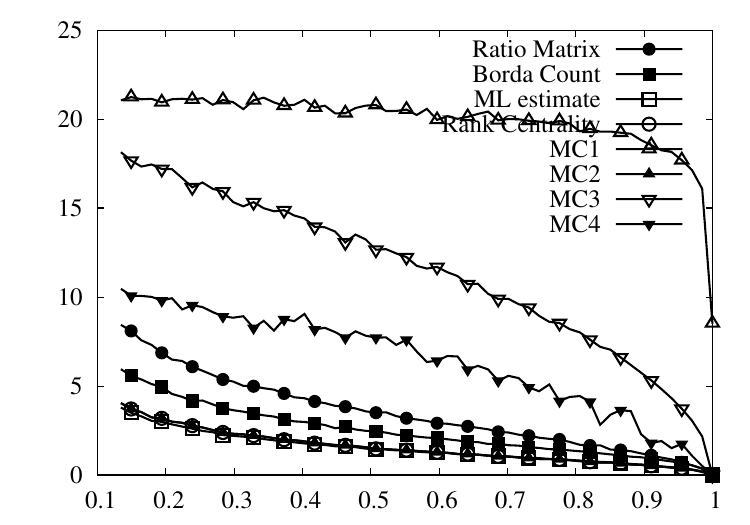}
	\put(-190,-5){edge sampling rate}
	\put(-380,150){$D_{L_1}(\sigma_{\rm GT},\sigma_{\rm Sample})$}
  \end{center}
  \caption{\small{{Experimental results on a real dataset shows that 
  Rank Centrality, ML estimator and MC2 are less sensitive to having limited data. } }}
  \label{fig:experiment2}
\end{figure}

Now each algorithm gives different ground truth rankings
\snncomment{given the full set of data}. This ground truth is compared
to a ranking we get from only a subset of the data, which is generated
by sampling each edge with a given sampling rate and revealing only
the data on those sampled edges.  We want to measure how much each
algorithm is 
\DS{affected} by eliminating edges from the complete graph.  Let
$\sigma_{\rm GT}$ be the ranking we get by applying our choice of rank
aggregation algorithm to the complete dataset, and $\sigma_{\rm
  Sample}$ be the ranking we get from sampled dataset.  To measure the
resulting error in the ranking, we use the following metric:
\begin{eqnarray*}
	D_{L_1}(\sigma_{\rm GT},\sigma_{\rm Sample}) &=& \frac{1}{\numitems}\sum_i|\sigma_{\rm GT}(i)-\sigma_{\rm Sample}(i)| \;.
\end{eqnarray*}
Figure~\ref{fig:experiment2} illustrates that 
\finalcomment{Rank Centrality, ML estimator and MC2 are less sensitive to sampling the dataset, compared to Borda Count, MC1, MC3, and MC4.  
Hence they are more robust when available comparisons data is limited. }

\medskip
\noindent {\em Dataset 2: NASCAR 2002.} Table \ref{tab:nascar} shows ranking of drivers from NASCAR 2002 season racing results. 
\citet{Hun04} used this dataset for studying rank-aggregation algorithms, and we use the dataset, publicly available at \citep{GS09}: \\
\texttt{http://sites.stat.psu.edu/$\sim$dhunter/code/btmatlab/}. \\
The dataset has 87 different drivers who competed in total 36 races in which 43 drivers were racing at each race. 
Some of the drivers raced in all 36 races, whereas some drivers only participated in one. To break the racing results into parities comparisons 
and to be able to run the comparison based algorithm, like \citet{Hun04,GS09}, we eliminated four drivers who finished last in every 
race they participated. Therefore, the dataset we used, there are total 83 drivers. 

Table \ref{tab:nascar} shows top ten and bottom ten drivers according to their average place, 
and their ranking from Rank Centrality and ML estimator. 
The unregularized Rank Centrality can over fit the data by placing P. J. Jones and Scott Pruett in the first and second places. 
They have high average place, but they only participated in one race. 
In contrast, the regularized version places them lower and gives the top ranking to those players with more races. 
Similarly, Morgan Shepherd is placed last in the regularized version, because he had 
consistently low performance in 5 races. \ourchanges{Similarly, the ML estimator  with regularization gives the top (and bottom) rankings to those players with more races.}

	  \begin{table}[ht]
	  \centering
	 { \scriptsize
		  \begin{tabular}{|c c c |c c |c c|c c |}
		  \hline
		  \multirow{3}{*}{Driver}& \multirow{3}{*}{Races} & \multirow{3}{*}{Av. place}  & \multicolumn{4}{|c|}{Rank Centrality}&
		 \multicolumn{2}{|c|}{ML estimator}\\ 
		 && &     \multicolumn{2}{|c|}{$\varepsilon=0$}&  \multicolumn{2}{|c|}{$\varepsilon=3$}&\multicolumn{2}{|c|}{$\lambda=0.01$}\\[0.5ex]
		 && &     $\pi$& rank & $\pi$ & rank & $e^\theta$ &rank \\[0.5ex]
		  \hline
		   	P. J. Jones & 1 &4.00&     			0.1837&1& 0.0181&11& 0.0124&23 \\ 
		   	Scott Pruett & 1 &4.00&     		0.0877&2& 0.0176&12& 0.0124& 24\\ 
		   	Mark Martin & 36 &12.17&     		0.0302&5& 0.0220&2& 0.0203& 1\\ 
		   	Tony Stewart & 36 &12.61&     		0.0485&3& 0.0219&1& 0.0199& 2\\ 
		   	Rusty Wallace & 36 &13.17&     	0.0271&6& 0.0209&3& 0.0193& 3\\ 
		   	Jimmie Johnson & 36 &13.50&     	0.0211&12& 0.0199&5& 0.0189& 4\\ 
		   	Sterling Marlin & 29 &13.86&     	0.0187&14& 0.0189&10& 0.0177& 8\\ 
		   	Mike Bliss & 1 &14.00&     		0.0225&10& 0.0148&18& 0.0121& 27\\ 
		   	Jeff Gordon & 36 &14.06&     		0.0196&13& 0.0193&8& 0.0184& 5\\ 
		   	Kurt Busch& 36 &14.06&     		0.0253&7& 0.0200&4& 0.0184& 6\\ 
		   	$\vdots$ &  & &     				&& && & \\ 
		   	Carl Long & 2 & 40.50 &     		0.0004&77& 0.0087&68& 0.0106& 59\\ 
		   	Christian Fittipaldi & 1 & 41.00&     	0.0001&83& 0.0105&49& 0.0111& 40\\ 
		   	Hideo Fukuyama & 2 & 41.00&     	0.0004&76& 0.0088&67& 0.0106& 60\\ 
		   	Jason Small & 1 & 41.00&     		0.0002&80& 0.0105&48& 0.0111& 41\\ 
		   	Morgan Shepherd & 5 & 41.20&     	0.0002&78& 0.0059&83& 0.0092& 75\\ 
		   	Kirk Shelmerdine & 2 & 41.50&     	0.0002&81& 0.0084&70& 0.0105& 61\\ 
		   	Austin Cameron & 1 & 42.00&     	0.0005&75& 0.0107&44& 0.0111& 43\\ 
		   	Dave Marcis & 1 & 42.00&     		0.0012&71& 0.0105&47& 0.0111& 44\\ 
		   	Dick Trickle & 3 & 42.00&     		0.0001&82& 0.0071&77& 0.0100& 65\\ 
		   	Joe Varde & 1 & 42.00&     		0.0002&79& 0.0110&43& 0.0111& 42\\ 
		  \hline
		  \end{tabular}
		  \caption{$\varepsilon$-regularized Rank Centrality for top ten and bottom ten 2002 NASCAR drivers, 
		  as ranked by average place. }
		  \label{tab:nascar}
	  }
	  \end{table}

\medskip
\noindent {\em Dataset 3: ODI Cricket.} Table \ref{tab:cricket} shows ranking of international cricket teams from the 2012 season of 
the One Day International (ODI) cricket match, where \ourchanges{16 teams played total of 362 games}. 
Like NASCAR dataset, in Table \ref{tab:cricket}, teams with smaller number of matches, such as 
Scotland and Ireland, are moved towards the middle with regularization, and New Zealand is moved towards the end. 
Notice that regularized or not, the ranking from Rank Centrality is different from the simple ranking from average place or 
winning ratio, because we give more score for winning against stronger opponents. \ourchanges{The regularized ML estimator produces similar ranking as the regularized Rank Centrality. } 
\finalcomment{This data on ODI cricket match is publicly available, for example from \texttt{http://www.cricmetric.com/blog/}.}

	  \begin{table}[ht]
	  \centering
	 { \scriptsize
		  \begin{tabular}{|c c c c |c c |c c|c c |}
		  \hline
		  \multirow{3}{*}{Team}& \multirow{3}{*}{matches} & \multirow{3}{*}{Win ratio} & \multirow{3}{*}{deg} & \multicolumn{4}{|c|}{Rank Centrality}&
		 \multicolumn{2}{|c|}{ML estimator}\\ 
		 &&& &     \multicolumn{2}{|c|}{$\varepsilon=0$}&  \multicolumn{2}{|c|}{$\varepsilon=1$}&\multicolumn{2}{|c|}{$\lambda=0.01$}\\[0.5ex]
		 &&& &     $\pi$& rank & $\pi$ & rank & $e^\theta$ &rank \\[0.5ex]
		  \hline
		  South Africa 	&43&0.6744&     		11&0.1794&2&0.0943&2&0.0924&2\\ 
		  India 		&76&0.6382&     		11&0.1317&4&0.0911&3&0.0923&3\\ 
		  Australia 	&72&0.6319&     		13&0.1798&1&0.0900&4&0.0881&4\\ 
		  England 		&60&0.6000& 		10&0.1526&3&0.0957&1&0.0927&1\\ 
		   Scotland 	&15&0.6000&		7&0.0029&12&0.0620&7&0.0627&7\\ 
		  Sri Lanka 	&78&0.5577&     		12&0.1243&5&0.0801&5&0.0768&5\\ 
		  Parkistan 	&65&0.5385&     		13&0.0762&6&0.0715&6&0.0755&6\\ 
		   Ireland 		&32&0.5316&    		13&0.0124&11&0.0561&8&0.0539&9\\ 
		   Afghanistan 	&20&0.5000& 		7&0.0005&15&0.0435&13&0.0472&12\\ 
		  West Indies 	&55&0.4091&     		12&0.0396&7&0.0546&9&0.0592&8\\ 
		  New Zealand 	&50&0.3800& 		10&0.0354&8&0.0466&12&0.0514&10\\ 
		  Bangladesh 	&51&0.3333&     		11&0.0320&9&0.0500&10&0.0492&11\\ 
		   Netherlands 	&24&0.3333& 		10&0.0017&13&0.0432&14& 0.0427&14\\ 
		   Zimbabwe 	&40&0.3250& 		11&0.0307&10&0.0481&11&0.0439&13\\ 
		   Canada 		&22&0.2273&		11& 0.0003&16&0.0365&16&0.0364&15\\
		   Kenya 		&21&0.1905& 		10&0.0007&14&0.0367&15&0.0356&16\\ 
		  \hline
		  \end{tabular}
		  \caption{Applying $\varepsilon$-regularized Rank Centrality  
		  to One Day International (ODI) cricket match results from 2012.
		  The degree of a team in the comparisons graph is the number of teams it has played against. 
		  }
		  \label{tab:cricket}
	  }
	  \end{table}

\subsection{Information-theoretic lower bound}
\label{sec:lowerbound}

In previous sections, we presented the achievable error rate based on a 
particular low-complexity algorithm. 
In this section, we ask how this bound compares to the fundamental limit 
under BTL model. 

Our result in Theorem~\ref{thm:mainER} provides an upper bound on 
the achievable error rate 
between estimated scores and the true underlying scores. 
We provide a constructive argument to lower bound the minimax error rate 
over a class of BTL models. 
Concretely, we consider the scores coming from a simplex with bounded dynamic range defined as  
\begin{eqnarray*}
	\cS_\upbound &\equiv&\Big\{\tpi\in\reals^{\numitems} \;\Big|\; \sum_{i\in[\numitems]}\tpi_i=1\;,\; \max_{i,j}\frac{\tpi_i}{\tpi_j}\leq \upbound \Big\}\;. 
\end{eqnarray*}
We constrain the scores to be on the simplex, because 
we represent the scores by its projection onto the standard simplex 
as explained in Section~\ref{sec:BTL}. 
Then, we can prove the following lower bound on the minimax error rate.
\begin{theorem}
	\label{thm:lowerbound}
	\DS{Consider a minimax scenario where we first choose an algorithm ${\cal A}$ that estimates the BTL 
	weights, say $\pi^{\cal A}$, from given observations and for this particular algorithm ${\cal A}$, nature 
	chooses the worst-case true BTL weights $\tpi$. Let  $\cS_\upbound$ denote the space of all BTL 
	score vectors $\tpi$ with dynamic range at most $b$ as defined above. Then
	\begin{eqnarray}
		\inf_{{\cal A}}\; \sup_{\tpi\in\cS_\upbound}\; \frac{\E\big[\,\|\pi^{\cA}-\tpi\|\,\big]}{\|\tpi\|} &\geq& \frac{\upbound-1}{240\sqrt{10}(\upbound+1)}\frac{1}{\sqrt{k\numcomp}}\;,
		\label{eq:lowerbound}
	\end{eqnarray}
	where the infimum ranges over all estimation algorithms $\cA$ that are 
	measurable functions over the observations. Here a pair of items is chosen to be compared with probability $\numcomp/\numitems$, and
	for thus chosen pair $k$ comparison observations are generated as per the underlying BTL model.}
\end{theorem}

By definition the dynamic range is always at least one. 
When $b=1$, we can trivially achieve a minimax rate of zero.
Since the infimum ranges over all measurable functions, 
it includes a trivial estimator which always outputs 
$(1/\numitems)\ones$ regardless of the observations,  
and this estimator achieves zero error when $b=1$. 
In the regime where the dynamic range $b$ is bounded away from one and bounded above by a constant, 
Theorem~\ref{thm:lowerbound} establishes that the upper bound obtained in Theorem~\ref{thm:mainER} is minimax-optimal 
up to factors logarithmic in the number of items $\numitems$. 

\subsection{{MLE: Error bounds using state-of-art method}}
\label{sec:ML}

It is well known that the maximum-likelihood estimate of a set of
parameters is asymptotically normal with mean $0$ and covariance equal
to the inverse Fisher information of the set of parameters. In this
section we wish to show the behavior of the estimates obtained through
the logistic regression based approach for estimating the parameters
$\thetaparam_i = \log \wparam_i$ in a finite sample setting.

\noindent{\bf Model.}
Recall that the logistic regression based method reparameterizes the model
so that given items $i$ and $j$ the probability that $i$ defeats $j$ is
\begin{equation*}
  P(i \textrm{ defeats } j) = \frac{\exp(\thetaparam_i - \thetaparam_j)}{1+\exp(\thetaparam_i - \thetaparam_j)}.
\end{equation*}
In order to ensure identifiability we also assume that $\sum_i
\thetaparam_i = 0$, so that we also enforce the constraint $\sum
\thetahat_i = 0$. We also recall that we let $b =
\wparam_{\max}/\wparam_{\min}$. Similarly, we let $\btilde \defn
\thetaparam_{\max} - \thetaparam_{\min}$ and enforce the constraint
that $\thetahat_{\max} - \thetahat_{\min} \leq \btilde^{'}$ where
$\btilde \leq \btilde^{'}$. For simplicity we assume that $\btilde^{'} =
\btilde$.

Finally, recall that we are given $\numobs$ i.i.d. observations. We
take $l \in \{1,2,\hdots,\numitems\}$ and let $\mleobs_l$ to be the
outcome of the $l^{\textrm{th}}$ comparison. Furthermore, if during
the $l^{\textrm{th}}$ competition item $i$ competed against item $j$
we take $\design_{l} = e_i - e_j$ where $e_i$ is the standard basis
vector with entries that are all zero except for the $i^{th}$ entry,
which equals one. Note that in this context the ordering of the
competition does matter. Finally, we define the inner-product between
two vectors $x,y \in \reals^{\numitems}$ to be $\inprod{x}{y} =
\sum_{i=1}^\numitems x_i y_i$. Therefore, under the BTL model with
parameters $\thetaparam$ we have that
\begin{equation*}
  \mleobs_l = \begin{cases}
    1 & \text{ with probability $\exp{\inprod{\design_l}{\thetaparam}}/(1+\exp{\inprod{\design_l}{\thetaparam}})$}\\
    0 & \text{ otherwise}.
    \end{cases}
\end{equation*}
Now the estimation procedure is of the form
\begin{equation*}
  \thetahat = \argmin_{\theta} \Loss(\theta,\mleobs,\design)
\end{equation*}
where
\begin{equation}
  \label{defn:loss}
  \Loss(\theta,\mleobs,\design) = \frac{1}{\numobs} \sum_{l=1}^n \log(1+\exp{\inprod{\design_l}{\theta}}) - \mleobs_l \inprod{\design_l}{\theta}
\end{equation}

\noindent{\bf Results.}
Before proceeding we recall that $\| \thetaparam
\|_2 \leq \btilde \sqrt{\numitems}$. With that in mind we have the
following theorem.
\begin{theorem}\label{thm:MLE}
  Suppose that we have $\numobs > 12 \numitems \log \numitems$
  observations of the form $(i,j,y)$ where $i$ and $j$ are drawn
  uniformly at random from $[\numitems]$ and $y$ is Bernoulli with
  parameter $\exp(\theta^*_i - \theta^*_j)/(1+\exp(\theta^*_i -
  \theta^*_j))$. Then, we have with probability at least $1-2/\numitems$
  \begin{equation*}
    \label{thm:mle}
    \|\thetahat - \thetaparam\| \; \leq\;  6 \frac{(1+b)^2}{b} \sqrt{\frac{\numitems^2 \log \numitems}{\numobs}}.
  \end{equation*}
\end{theorem}
With the assumption that $\|\thetaparam \|_\infty \leq \btilde$, we
have $\|\thetaparam\| \leq \btilde \sqrt{\numitems}$.

\subsection{{Cram\'er-Rao lower bound}}
\label{subsec:cramer}
The Fisher information matrix (FIM) encodes the amount of information that the observed measurements 
carry about the parameter of interest. 
The Cram\'er-Rao bounds we derive in this section provides a lower bound on the expected squared Euclidean norm $\E[\|\tpi-\pi\|^2]$ of any unbiased estimator and 
is directly related to the (inverse of) 
Fisher information matrix. 

Denote the log-likelihood function as  
\begin{eqnarray*}
	\ell(\tpi|a) &=& \sum_{(i,j)\in E} \log f(a_{ij},a_{ji}|\tpi) \;, \text{ where } \\
	f(a_{ij},a_{ji}|\tpi) &=& \Big(\frac{\tpi_j}{\tpi_i+\tpi_j}\Big)^{k_{ij} a_{ij}}\Big(\frac{\tpi_i}{\tpi_i+\tpi_j}\Big)^{k_{ij} a_{ji}} \;, 
\end{eqnarray*}
and $k_{ij}$ is the number of times the pair $(i,j)$ was compared. 
The Fisher information matrix with the BTL weights $\tpi$ is defined as $F(\tpi)\in\reals^{n\times n}$ with 
\begin{eqnarray*}
	F(\tpi)_{ij} &=& \E_a\Big[ -  \frac{\partial^2 \ell(\tpi|a)}{\partial \tpi_i \partial \tpi_j} \Big] \;=\; 
		  \left\{ \begin{array}{rl}
	                       \sum_{i'\in\partial i}\frac{k_{ii'}}{(\tpi_i+\tpi_{i'})^2}\frac{\tpi_{i'}}{\tpi_{i}}& \text{ if }i=j\;,\\
	                       -\frac{k_{ij}}{(\tpi_i+\tpi_j)^2}& \text{ if }(i,j)\in E\;,\\
	                       0& \text{ otherwise }\;.\\
	                      \end{array}
	              \right.
\end{eqnarray*}
This follows from the fact that 
\begin{eqnarray*}
	\frac{\partial \ell(\tpi|a)}{\partial \tpi_i  } &=& \sum_{i'\in\partial i} \frac{-k_{ii'} (a_{ii'}+a_{i'i})}{\tpi_i+\tpi_{i'}} + \frac{k_{ii'}a_{i'i}}{\tpi_i}\;, \text{ and }\\
	\frac{\partial^2 \ell(\tpi|a)}{\partial \tpi_i \partial\tpi_{j}  } &=& \left\{ 
	\begin{array}{rl}
	\sum_{i'\in\partial i} k_{ii'}\Big( \frac{1}{(\tpi_i+\tpi_{i'})^2} - \frac{a_{i'i}}{(\tpi_i)^2}\Big)
	&\text{ if $i=j$}\;,\\
	    \frac{k_{ij}}{(\tpi_i+\tpi_{j})^2} 
	&\text{ if $(i,j)\in E$}\;,\\
	0&\text{ otherwise}\;.
	\end{array}\right.
\end{eqnarray*}
Let $\pi$ denote our estimate of the weights. 
Applying the Cram\'er-Rao bound \citep{Rao45}, we get the following lower bound for all unbiased estimators $\pi$:  
\begin{eqnarray*}
	E[\|\pi-\tpi\|^2] & \geq &{\rm Trace}(F(\tpi)^{-1})
\end{eqnarray*}
This bound depends on $\tpi$ and the graph structure. 
Although a closed form expression is difficult to get and Rank Centrality as well as the ML estimate is biased, 
we compare our numerical experiments with a numerically computed Cram\'er-Rao bound on the same graph and the same weights $\tpi$. 

\subsubsection{{Numerical comparisons}}

\begin{figure}
\begin{center}
\includegraphics[width=.3\textwidth]{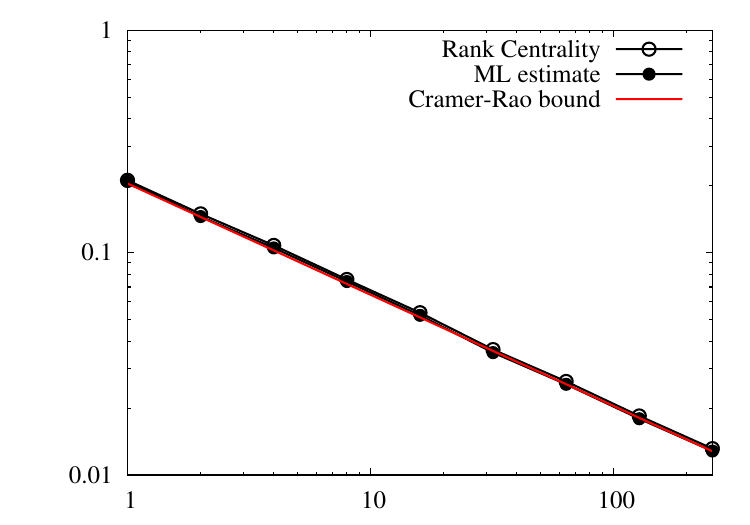}
\put(-155,65){\footnotesize RMSE}
\put(-60,-7){\footnotesize$k$}
\includegraphics[width=.3\textwidth]{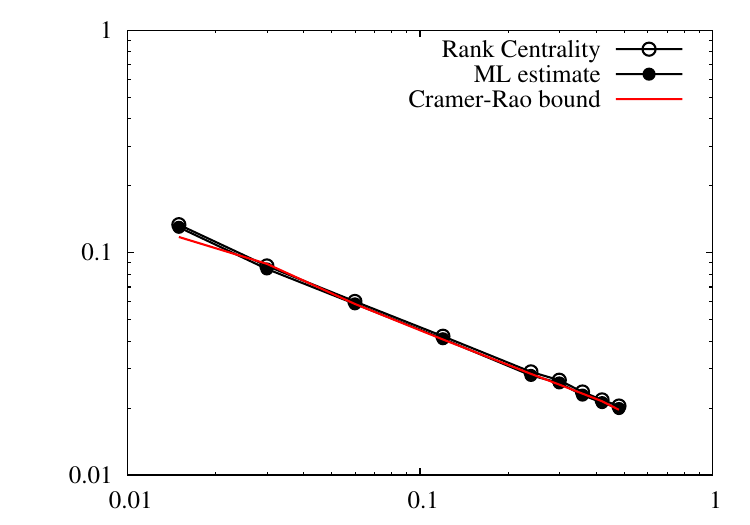}
\put(-60,-7){\footnotesize$d/n$}
\includegraphics[width=.3\textwidth]{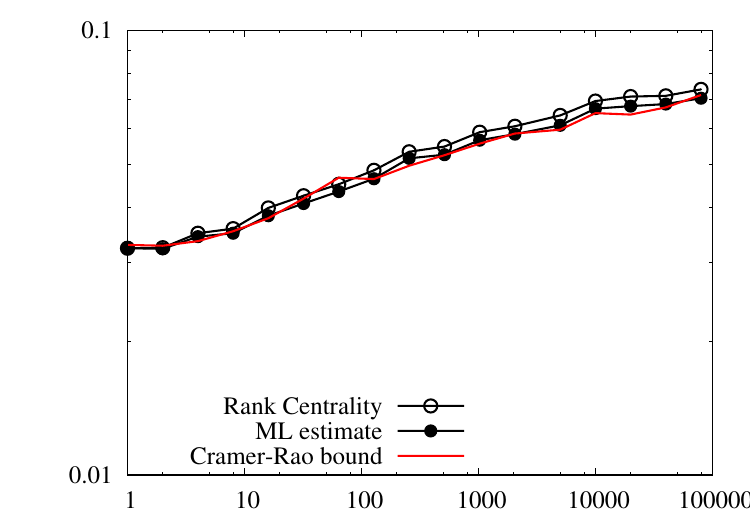}
\put(-60,-7){\footnotesize$b$}
\end{center}
\caption{Comparisons of Rank Centrality, the ML estimator, and the Cram\'er-Rao bound. All three lines are almost indistinguishable for all ranges of model parameters. }
\label{fig:crb}
\end{figure}

In Figure \ref{fig:crb}, 
the average normalized root mean squared error (RMSE) is shown as a function of various model parameters. We fixed the control parameters as $k=32$, $n=400$, $d=60$ and $b=10$ with pairs assigned according to Erd\"os-Renyi graph $G(n,d/n)$.
Each point in the figure is averaged over $20$ random instances ${\cal S}$. 
Let $\tpi^{(i)}$ be the resulting estimate at $i$-th experiment, then   
\begin{eqnarray}
	\label{eq:rmse}
	{\rm RMSE} &=& \frac{1}{|{\cal S}|} \sum_{i\in{\cal S}} \frac{\|\pi^{(i)}-\tpi\| }{\|\tpi\|}
\end{eqnarray}
For all ranges of model parameters $k$, $d$, and $b$, 
RMSE achieved using Rank Centrality is almost indistinguishable from 
that of the ML estimate and also the Cram\'er-Rao bound (CRB). 

{CRB provides a lower bound on the expected mean squared error for unbiased estimators. 
Although we are plotting average root mean squared error, as opposed to 
average mean squared error, 
we do not expect any estimator to achieve RMSE better than the CRB 
as long as there is a concentration. }

{The ML estimator in \eqref{eq:logit} with $\lambda=0$ finds an estimate $\pi=e^{\hat{\theta}}$ 
that maximizes the log-likelihood, 
and in general ML estimate does not coincide with the minimum mean squared error estimator. 
From the figure we see that it intact achieves the minimum mean squared error and matches the CRB. 
}

{What is perhaps surprising is that for all the parameters that we experimented with, 
the RMSE achieved by Rank Centrality is almost indistinguishable with that of ML estimate and the CRB. 
Thus, coupled with the minimax lower-bounds, one cannot do better than Rank Centrality \finalcomment{under the BTL model.} 
}

%
%
\subsection{Discussion of Results}
\finalcomment{ In this section we review the results that we have
  established above. In Theorem~\ref{thm:maingeneral} we establish
  upper bounds on the error when samples are drawn from an arbitrary
  graph and when each edge is compared $k$ times. This bound depends
  on the spectral gap of the underlying graph, which shows that graphs
  with a larger spectral gap achieve smaller estimation error.  For
  the case of Erd\"os-Renyi graphs, Theorem \ref{thm:mainER} provides
  an upper bound on the error achieved by Rank Centrality. In
  Theorem~\ref{thm:lowerbound} we prove that the bound is
  near-optimal, up to logarithmic factors, in an information theoretic
  sense. That is, no method, regardless of computational power can achieve better performance
  on the same statistical model. For a tighter analysis of the
  optimality of Rank Centrality, we provide numerical experiments
  under the BTL model and compare it to the Cramer Rao lower-bound established in Section~\ref{subsec:cramer}.
  Comparisons with the Cramer-Rao bound in Figure \ref{fig:crb}
  suggests that the error achieved by Rank Centrality is
  indistinguishable from the fundamental Cramer-Rao lower bound, and
  hence exactly optimal for a certain class of estimators.  }

\finalcomment{ For completeness, we further provide an analysis of the
  error achieved by the MLE in Theorem~\ref{thm:MLE}. Building upon our 
  analysis, \cite{HOX14} shows that MLE is near order-optimal, just like
  \RC. }

\finalcomment{Finally, we compare the computational cost of Rank Centrality 
versus the MLE. While it is difficult to make an exact, theoretical, comparison, 
  we nevertheless compare their computational cost by means of popular implementations
  on a common computation platform. For \RC, the implementation is based
  on using {\em eigs} function MATLAB. For MLE, the implementation is based
  on the basic first-order method. In a collection of experiments (with varying 
  problem parameters), \RC~converges an order of magnitude faster than the
  MLE. It should be noted that the first-order method has tunable step-size and 
  our implementation did not attempt to optimize this selection when varying 
  problem parameters. Finally, MLE can be viewed as a standard logistic regression. 
  Therefore, the {\em lm} function of R-package can be used to solve for MLE.
  Again, in the same computation environment, the resulting MLE is order of
  magnitude slower compared to the MATLAB implementation of \RC, but faster than the first-order method.}
  
%
%
\section{Proofs}
\label{SecProofs}
We may now present proofs of Theorems~\ref{thm:maingeneral} and \ref{thm:mainER}. 
We first present a proof of convergence for general graphs 
in Theorem~\ref{thm:maingeneral}. 
This result follows from Lemma \ref{lem:perturbation} that we state below, which shows that our algorithm
enjoys convergence properties that result in useful upper bounds. 
The lemma is made general and uses standard techniques of spectral
theory. 
The main difficulty arises in establishing that the Markov
chain $P$ satisfies certain properties that we will discuss subsequently. Given the proof for the general graph, 
Theorem~\ref{thm:mainER} follows by showing that in the case of Erd\"os-Renyi graphs, 
certain spectral properties are satisfied with high probability. 

The next set of proofs involve the information-theoretic lower bound stated in Theorem \ref{thm:lowerbound} and the 
proof of Theorem \ref{thm:MLE} establishing the finite sample error analysis of MLE. 

\subsection{Proof of Theorem \ref{thm:maingeneral}: General graph}
\label{sec:maingeneralproof}
In this section, we characterize the error rate achieved by our
ranking algorithm. Given the random Markov chain $P$, where the
randomness comes from the outcome of the comparisons, we will show
that it does not deviate too much from its expectation $\tP$, where we
recall that $\tP$ is defined as
\begin{eqnarray*}
    \label{eq:deftP}
    \tP_{ij}&=&\left\{
        \begin{array}{rl} 
	    \frac1\dmax \frac{w_j}{w_i+w_j} &\text{ if }i\neq j\;,\\
	    1-\frac{1}{\dmax}\sum_{\ell \neq i} \frac{w_\ell}{w_i+w_\ell}&\text{ if }i=j\;
	\end{array}
    \right.
\end{eqnarray*}
for all $(i,j) \in E$ and $\tP_{ij} = 0$ otherwise.

Recall from the discussion following equation~\eqref{eq:defP} that the
transition matrix $P$ used in our ranking algorithm has been carefully
chosen such that the corresponding expected transition matrix $\tP$
has two important properties.  First, the stationary distribution of
$\tP$, which we denote with $\tpi$ is proportional to the weight
vectors $\score$. Furthermore, when the graph is connected and has
self loops (which at least one exists), this Markov chain is
irreducible and aperiodic so that the stationary distribution is
unique. The next important property of $\tP$ is that it is
reversible--$\tpi(i)\tP_{ij}=\tpi(j)\tP_{ji}$. This observation
implies that the operator $\tP$ is symmetric in an appropriately defined
inner product space.
The symmetry of the operator $\tP$ will be crucial in applying
ideas from spectral analysis to prove our main results.

Let $\Delta$ denote the fluctuation of the transition matrix around
its mean, 
such that $\Delta \equiv P-\tP$.
The following lemma bounds the deviation of the Markov chain after $t$ steps 
in terms of two important quantities: 
the spectral radius of the fluctuation $\|\Delta\|_2$ and 
the spectral gap $1-\lmax(\tP)$, where 
\begin{eqnarray*}
  \lmax(\tP)
  &\equiv& \max\{\lambda_2(\tP),-\lambda_n(\tP)\}\;.
\end{eqnarray*}
Since $\lambda(\tP)$'s are sorted, 
$\lmax(\tP)$ is the second largest eigenvalue in absolute value.
\begin{lemma}
    \label{lem:perturbation}
    For any Markov chain $P=\tP+\Delta$ with a reversible Markov chain $\tP$, 
    let $p_t$ be the distribution of the Markov chain $P$ when 
    started with initial distribution $p_0$. Then, 
    \begin{eqnarray}
      \frac{\big\|p_t-\tpi\big\|}{\|\tpi\|}&\leq&\rho^t\frac{\|p_0-\tpi\|}{\|\tpi\|}\sqrt{\frac{\tpi_{\rm max}}{\tpi_{\rm min}}} + \frac{1}{1-\rho}\|\Delta\|_2\sqrt{\frac{\tpi_{\rm max}}{\tpi_{\rm min}}}\;.
      \label{eq:perturbation}
    \end{eqnarray}
    where $\tpi$ is the stationary distribution of $\tP$, 
     $\tpi_{\rm min}=\min_i\tpi(i)$, $\tpi_{\rm max}=\max_i\tpi(i)$, 
     and $\rho=\lmax(\tP)+\|\Delta\|_2\sqrt{\tpi_{\rm max}/\tpi_{\rm min}}$.
\end{lemma}
The above result provides a general mechanism for establishing error
bounds between an estimated stationary distribution $\pi$ and the
desired stationary distribution $\tpi$. It is worth noting that the
result only requires control on the quantities $\|\Delta\|_2$ and $1 -
\rho$. We may now state two technical lemmas that provide control on
the quantities $\|\Delta\|_2$ and $1-\rho$, respectively.
\begin{lemma}
  \label{lem:boundDel}
  For \SN{some constant $C \geq 8$,
  the error matrix $\Delta = P - \tP$ satisfies
  \begin{eqnarray*}
    \|\Delta\|_2 &\leq& C \sqrt{\frac{\log \numitems}{\resamp\,\dmax}}
  \end{eqnarray*}
  with probability at least $1-4 \numitems^{-C/8}$.}
 \end{lemma}
The next lemma provides our desired bound on $1-\rho$.
\begin{lemma}
  \label{lem:specGap}
  \SN{If} $\|\Delta\|_2\leq C\sqrt{\log n/(k \dmax)}$ 
  and $k\geq 4C^2b^5\dmax\log n (1/\dmin\eiggap)^2$, 
  \snncomment{then}
  \begin{eqnarray*}
    1-\rho &\geq& \frac{\eiggap\dmin}{b^2\dmax}\; .
  \end{eqnarray*}
\end{lemma}

\noindent
{\em Proof of Theorem~\ref{thm:maingeneral}.}  With the above stated Lemmas,  we shall proceed with the 
proof of Theorem~\ref{thm:maingeneral}. When there is a positive spectral gap such that $\rho<1$, 
the first term in \eqref{eq:perturbation} vanishes
as $t$ grows.  The rest of the first term is bounded and independent
of $t$.  Formally, we have
$$\tpi_{\rm max}/\tpi_{\rm min}\leq b\;,\;\; \|\tpi\|\geq1/\sqrt{n}\;,\;\; \text{ and }\|p_0-\tpi\|\leq2\;,$$
by the assumption that $\max_{i,j} w_i/w_j \leq b$ and the fact that
$\tpi(i)=w_i/(\sum_j w_j)$.  Hence, the error between the distribution
at the $t^{th}$ iteration $p^t$ and the true stationary distribution
$\tpi$ is dominated by the second term in
equation~\eqref{eq:perturbation}. 
Substituting the bounds in Lemma~\ref{lem:boundDel} 
and Lemma~\ref{lem:specGap}, the dominant second
term in equation~\eqref{eq:perturbation} is bounded by
\begin{eqnarray*}
	\lim_{t\to\infty}\frac{\big\|p_t-\tpi\big\|}{\|\tpi\|}&\leq& 
	\frac{C\, b^{5/2}}{\eiggap\dmin}\, \sqrt{\frac{\dmax\log n}{k}}
\end{eqnarray*}
\SN{with probability at least $1 - 4 n^{-C/8}$.}
In fact, we only need 
$t=\Omega(\log n + \log b + \log(\dmax \log \numitems / (\dmin^2 k \eiggap^2)))$ to ensure that 
the above bound holds up to a constant factor. 
This finishes the proof of Theorem \ref{thm:maingeneral}.
Notice that in order for this result to hold, 
we need $k\geq 4C^2 b^5 \dmax\log n (1/\dmin \xi)^2$ for Lemma~\ref{lem:specGap}.


\subsubsection{Proof of Lemma \ref{lem:perturbation}.}
\label{sec:proofperturbation}

Due to the reversibility of $\tP$, we can view it as a self-adjoint
operator on an appropriately defined inner product space. This
observation allows us to apply the well-understood spectral analysis
of self-adjoint operators.  To that end, define an inner product space $L^2(\tpi)$ 
as a space of $n$-dimensional vectors, $\reals^n$, endowed with
\begin{eqnarray*} 
    \<a,b\>_\tpi=\sum_{i=1}^na_i\tpi_ib_i\;. 
\end{eqnarray*}
Similarly, we define $\|a\|_\tpi=\sqrt{\<a,a\>}_\tpi$ as the $2$-norm
in $L^2(\tpi)$.  \DS{An operator (matrix) $A$ is self-adjoint with respect to $L^2(\pi)$
if $\langle u, Av \rangle_\tpi = \langle Au, v\rangle_\tpi$ for all $u, v \in \reals^n$.}
For a self-adjoint operator $A$ in $L^2(\tpi)$, we
define $\|A\|_{\tpi,2}=\max_{a}{\|Aa\|_\tpi}/{\|a\|_\tpi}$ as the
operator norm.  These norms are related to the corresponding norms in
the Euclidean space through the following inequalities.
\begin{eqnarray}
    \sqrt{\tpi_{\rm min}}\,\|a\| \;\;\leq& \|a\|_\tpi &\leq\;\; \sqrt{\tpi_{\rm max}}\,\|a\|\;,\label{eq:bound2norm}\\
    \sqrt{\frac{\tpi_{\rm min}}{\tpi_{\rm max}}}\,\|A\|_2 \;\;\leq& \|A\|_{\tpi,2} &\leq\;\; \sqrt{\frac{\tpi_{\rm max}}{\tpi_{\rm min}}}\,\|A\|_2\;.\label{eq:boundopnorm}
\end{eqnarray}
\DS{It is easy to check that, a reversible Markov chain $\tP$ is self-adjoint in $L^2(\tpi)$ due to the 
{\em detailed-balanced condition}, where $\tpi$ is the unique stationary distribution of $\tP$. }

\DS{Consider symmetrized version of $\tP$, defined as $S =
\tPi^{1/2}\tP\tPi^{-1/2}$, where $\tPi$ is a diagonal matrix with
$\tPi_{ii}=\tpi(i)$.  Again, reversibility of $\tP$ makes $S$ symmetric.}
\DS{It can be verified that $\tP$ and $S$ have the same set of eigenvalues.  
By Perron-Frobenius theorem, the eigenvalues are in $[-1,1]$ with largest being equal to $1$. 
Let they be denoted as $1=\lambda_1\geq\lambda_2\geq\ldots\geq\lambda_n\geq-1$, and let
$\lmax = \max\{|\lambda_n|, \lambda_2\}$. Let
$u_i$ be the left eigenvector of $S$ corresponding to $\lambda_i$ for $1\leq i\leq n$. 
 Then the $i$th left eigenvector of $\tP$ is given by $v_i=\tPi^{1/2}u_i$.  Since the first left 
 eigenvector of $\tP$ is the stationary distribution, i.e. $v_1=\tpi$, we have that 
 $u_1(i)=\tpi(i)^{1/2}$ or $\tPi^{-1/2}u_1 = \ones$. Finally, define rank-1 projection of $S$ as $S_1 = \lambda_1 u_1 u_1^T = u_1 u_1^T$ 
 and let $\tP_1 =  \tPi^{-1/2}S_1\tPi^{1/2}$. }

\DS{Our interest is in Markov chain $P=\tP+\Delta$ and  iterates obtained from it $p_t^T=p_{t-1}^TP$. Then, }
\begin{eqnarray}
    p_t^T-{\tpi}^T&=&(p_{t-1}-\tpi)^T(\tP+\Delta)+{\tpi}^T\Delta\;.\label{eq:recursion}
\end{eqnarray}
\DS{Using the fact that $(p_\ell-\tpi)^T\tPi^{-1/2}u_1=(p_\ell-\tpi)^T\ones=0$ for any probability 
distribution $p_\ell$, we get $(p_\ell-\tpi)^T\tP_1=(p_\ell-\tpi)^T\tPi^{-1/2}u_1\lambda_1u_1^T\tPi^{1/2}=0$. 
Then, from \eqref{eq:recursion} we get }
\begin{eqnarray*}
    p_t^T-\tpi^T&=&(p_{t-1}-\tpi)^T(\tP-\tP_1+\Delta)+\tpi^T\Delta\;.
\end{eqnarray*}
By definition of $\tP_1$, it follows that $\|\tP-\tP_1\|_{\tpi,2}=\|S-S_1\|_2=\lmax$.
Let $\rho=\lmax+\|\Delta\|_{\tpi,2}$, then  
\begin{eqnarray*}
     \|p_t-\tpi\|_\tpi&\leq&\|p_{t-1}-\tpi\|_{\tpi}(\|\tP-\tP_1\|_{\tpi,2}+\|\Delta\|_{\tpi,2})+\|\tpi^T\Delta\|_{\tpi}\\
     &\leq&\rho^t\|p_0-\tpi\|_{\tpi} + \sum_{\ell=0}^{t-1}\rho^{t-1-\ell}\|\tpi^T\Delta\|_\tpi\;.
\end{eqnarray*}
Dividing each side by $\|\tpi\|$ and applying the bounds in \eqref{eq:bound2norm} and
\eqref{eq:boundopnorm}, we get 
\begin{eqnarray*}
     \frac{\|p_t-\tpi\|}{\|\tpi\|} &\leq& \rho^t\sqrt{\frac{\tpi_{\rm max}}{\tpi_{\rm min}}}\frac{\|p_0-\tpi\|}{\|\tpi\|} + \sum_{\ell=0}^{t-1}\rho^{t-1-\ell}\sqrt{\frac{\tpi_{\rm max}}{\tpi_{\rm min}}}\frac{\|\tpi^T\Delta\|}{\|\tpi\|}\;.
\end{eqnarray*}
This finishes the proof of the desired claim. 

\subsubsection{Proof of Lemma \ref{lem:boundDel}.}
\label{sec:boundDel}

Our interest is in bounding $\|\Delta\|_2$. Now $\Delta = P - \tP$ so that for $1 \leq i,j \leq n$, 
\begin{align}\label{eq:f1}
\Delta_{ij} & = \frac{1}{k\dmax} C_{ij}, 
\end{align}
where $C_{ij}$ is distributed as per $B(k, p_{ij}) - k p_{ij}$ if $(i,j) \in E$ and $C_{ij} = 0$ otherwise. Here $B(k,p_{ij})$ is a Binomial random variable
with parameter $k$ and $p_{ij} \equiv \frac{w_j}{w_i + w_j}$. It should be noted that $C_{ij} + C_{ji} = 0$ and $C_{ij}$ are independent across all 
the pairs with $i < j$. For $1\leq i\leq n$
\begin{align}
\Delta_{ii} & = P_{ii} - \tP_{ii} ~=~ \big(1-\sum_{j\neq i} P_{ij}\big) - \big(1-\sum_{j \neq i} \tP_{ij}\big) 
~= \sum_{j\neq i} \tP_{ij} - P_{ij} ~ = - \sum_{j\neq i} \Delta_{ij}. \label{eq:f2}
\end{align}
Given the above dependence between diagonal and off-diagonal entries, we shall bound $\|\Delta\|_2$ as follows: let $D$ be the diagonal matrix
with $D_{ii} = \Delta_{ii}$ for $1\leq i\leq n$ and $\bar{\Delta} = \Delta - D$. Then, 
\begin{align}
\|\Delta\|_2 & = \| D + \bar{\Delta}\|_2 ~\leq \|D\|_2 + \|\bar{\Delta}\|_2. \label{eq:f3}
\end{align}
We shall establish the bound of $O\big(\sqrt{\frac{\log n}{k\dmax}}\big)$ for both $\|D\|_2$ and $\|\bar{\Delta}\|_2$ to establish the Lemma \ref{lem:boundDel}. 

\medskip
\noindent {\em Bounding $\|D\|_2$.} Since $D$ is a diagonal matrix, $\|D\|_2 = \max_{i} |D_{ii}| = \max_i |\Delta_{ii}|$. For a given fixed $i$, as per \eqref{eq:f1}-\eqref{eq:f2}, 
$k \dmax \Delta_{ii}$ can be expressed as summation of at most $k\dmax$ independent, zero-mean random variables taking values in the range of at most $1$. Therefore, by an application of Azuma-Hoeffding's inequality, it follows that 
\begin{align}
\prob\big(k\dmax |\Delta_{ii}| > t \big) & \leq 2 \exp\big(-\frac{t^2}{2 k\dmax}\big). 
\end{align}
By selection of $t = C \sqrt{k \dmax \log n}$ for appropriately large constant, it follows from above display that  
\begin{align}
  \prob \left ( \|D\|_2 \geq C \sqrt{\frac{\log n}{k \dmax}}  \right ) & \leq \sum_{i=1}^n \prob\Big(|\Delta_{ii}| > C \sqrt{\frac{\log n}{k\dmax}} \Big) \\
  & \leq 2 n^{-C^2/2 + 1}
\end{align}
\medskip
\noindent{\em Bounding $\|\bar{\Delta}\|_2$ when $\dmax \leq \log n$.} Towards this goal, we shall 
make use of the following standard inequality: for any square matrix $M$, 
\begin{align}\label{eq:normineq}
\|M\|_2 & \leq \sqrt{\|M\|_1 \|M\|_{\infty}},
\end{align} 
\DS{where $\|M\|_1 = \max_{i} \sum_{j} |M_{ij}|$ and $\|M\|_\infty = \|M^T\|_1$. In words, $\|M\|_2^2$ is
bounded above by product of the maximal row-sum and column-sum of absolute values of $M$. Since 
$\Delta_{ij}$ and $\Delta_{ji}$ are identically distributed and entries along each row 
(and hence each column) are independent, it is sufficient to obtain a high probability bound ($\geq 1-1/{\sf poly}(n)$) 
for maximal row-sum of absolute values of $\bar{\Delta}$; exactly the same bound will apply for column-sum using; and
using union bound the desired result will follow. }

To that end, consider the sum of the absolute values of the $i$th row-sum of $\bar{\Delta}$ and for simplicity let us denote
it by $R_i$. Then, 
\begin{align}
R_i & = \frac{1}{k\dmax} \sum_{j \neq i} |C_{ij}|, 
\end{align}
where recall that $C_{ij} = X_{ij} - k p_{ij}$ with $X_{ij}$ an independent Binomial random variable with parameters $k, p_{ij}$. Therefore,
for any $s > 0$,
\comment{
\begin{align}
\prob\big(R_i > s \big) & = \prob\big(\sum_j |C_{ij}| > k \dmax s\big) \nonumber \\
& \leq \E[\exp(\sum_j \theta |C_{ij}|)] \exp(-\theta k \dmax s), \quad \text{for any}~\theta > 0, \nonumber \\
& = \exp(-\theta k \dmax s) \prod_j \E[\exp(\theta |C_{ij}|)]. \label{eq:rc.1} 
\end{align}
}
\SN{
  \begin{align*}
    \prob\big(R_i > s \big) & = \prob\big(\sum_{j \in \partial i} |C_{ij}| > k \dmax s\big) \\
  & \leq \sum_{j \in \partial i} \sum_{\xi_j \in \{-1,+1\}} \prob \big (\sum_j \xi_j C_{i,j} > k \dmax s \big) \quad \text{by the union bound} \\
  & \leq \sum_{j \in \partial i} \sum_{\xi_j \in \{-1,+1\}} \exp \left ( \frac{-2 k^2 \dmax^2 s^2}{d_i k} \right )
  \end{align*}
  where the last inequality follows from Hoeffding's bound and the fact that
  $X_{ij} = \sum_{j=1}^k (y_{ij} - p_{ij})$ where $y_{ij}$ are Bernoulli random variables with mean $p_{ij}$. Now, the number of terms
  in the sum is $2^{d_i}$, the summand is constant, and $d_i \leq \dmax$. Thus, the last inequality is upper-bounded by
  \begin{align*}
    \sum_{j \in \partial i} \sum_{\xi_j \in \{-1,+1\}} \exp \left ( \frac{-2 k^2 \dmax^2 s^2}{d_i k^2} \right ) & \leq
    \exp \left ( -2 k \dmax s^2 + d_i \ln 2 \right )
  \end{align*}
}
By an application of the union bound
\begin{align*}
  \prob \left ( \| \barDelta \|_2 \geq s \right ) & \leq 2 n \prob \left (R_i \geq s \right ) \\
  & \leq 2 n \exp \left ( -2 k \dmax s^2 + \dmax \ln 2 \right ).
\end{align*}
Now, if we set $s = \frac{C}{2} \sqrt{\frac{\log n + \dmax \ln 2}{k
    \dmax}}$ we have that
\begin{equation*}
  \prob \left ( \| \barDelta \|_2 \geq C/2 \sqrt{\frac{\log n + \dmax \ln 2}{k \dmax}} \right ) \leq
    2 n^{-(C^2/2 - 1)}
\end{equation*}
Finally, using the assumption that $\dmax \leq \log n$ yields
\begin{equation*}
  \| \barDelta \|_2 \leq C \sqrt{\frac{\log n}{k \dmax}}
\end{equation*}
with probability at least $1 - 2 n^{-C^2/2+1}$.

\medskip
\noindent{\em Bounding $\|\bar{\Delta}\|_2$ when $\dmax \geq \log n$.} Towards this goal, we shall 
make use of the recent results on the concentration of \SN{the} sum of independent 
random matrices. For completeness, we recall the following result \citep{Jo11}.
\begin{lemma}[Theorem 6.2~\citep{Jo11}]
    \label{lem:matrixbennet}
    Consider a finite sequence $\{\tZ^{ij}\}_{i<j}$ of independent
    random self-adjoint matrices with dimensions $n\times n$.  Assume that
    \begin{eqnarray*}
      \E[\tZ^{ij}]=0 \quad \text{ and } \quad \E (\tZ^{ij})^p \preceq \frac{p!}{2} \, R^{p-2} (\tA^{ij})^2
      \quad, \text{ for $p=2,3,4,\hdots$}
    \end{eqnarray*}
    Define $\tsigma^2 \equiv \| \sum_{i>j} (\tA^{ij})^2\|_2$.
    Then, for all $t\geq0$, 
    \begin{eqnarray*}
      \prob \Big (\,\Big\|\sum_{i<j}\tZ^{ij}\Big\|_2\geq t\,\Big)&\leq&2n\exp\Big\{\frac{-t^2/2}{\tsigma^2+Rt}\Big\}\;.
    \end{eqnarray*}
\end{lemma}
We wish to prove concentration results on $\bar{\Delta} = \Delta-D = \sum_{i<j}
Z^{ij}$ where 
\begin{eqnarray*}
	Z^{ij} &=& (e_i e_j^T - e_j e_i^T) (P_{ij} - \tP_{ij})\;\;\;\; \text{ for $(i,j)\in E$}\;,  
\end{eqnarray*}
and $Z^{ij}=0$ if $i$ and $j$ are not connected. 
The $Z^{ij}$'s as defined are zero-mean
and independent, however, they are not self-adjoint. 
Nevertheless, we can symmetrize it by applying 
the dilation ideas presented in the paper~\citep{Jo11}:
\begin{equation*}
  \tZ^{ij} \equiv \left (
    \begin{matrix}
    0 & Z^{ij} \\
    (Z^{ij})^T & 0
    \end{matrix}
  \right ).
\end{equation*}
Now we can apply the above lemma to these self-adjoint, independent and 
zero-mean random matrices. 

To find $R$ and $\tA^{ij}$'s that satisfy the conditions of the lemma, 
first consider a set of matrices $\{A^{ij}\}_{i<j}$ such that 
$\tZ^{ij}=\Delta_{ij}A^{ij} $ and 
\begin{equation*}
  A^{ij} \; = \; \left ( \begin{matrix}
      0 & e_i e_j^T - e_j e_i^T \\
      e_j e_i^T - e_i e_j^T & 0
      \end{matrix} \right )\;,
\end{equation*}
if $(i,j)\in E$ and zero otherwise.
In the following, we show that the condition on $p$-th moment is satisfied with 
$R=1/\sqrt{\resamp\dmax^2}$ and $(\tA^{ij})^2=(1/(\resamp\dmax^2))(A^{ij})^2$ such that  
\begin{eqnarray}
	\label{eq:matrixbernsteincondition}
	\E\big[(\tZ^{ij})^p\big] &\preceq& \frac{p!}{2} 
	\Big(\frac{1}{\sqrt{\resamp\dmax^2}}\Big)^{p-2}\frac{1}{\resamp\dmax^2}(A^{ij})^2 \;. 
\end{eqnarray}
We can also show that $\tsigma^2 \equiv \|\sum_{i<j}(\tA^{ij})^2\|_2= 1/(\resamp\dmax)$, 
since 
\begin{eqnarray*}
	\sum_{i<j}(\tA^{ij})^2 \;=\; \sum_{i<j} \frac{1}{\resamp\dmax^2}\ind_{((i,j)\in E)}
	\left ( \begin{matrix}
      e_ie_i^T+e_je_j^T & 0\\
      0 & e_ie_i^T+e_je_j^T 
      \end{matrix} \right )
	\;=\; \frac{1}{\resamp\dmax^2} \sum_{i=1}^{\numitems} d_i 
    \left ( \begin{matrix}
      e_i e_i^T & 0\\
      0 & e_i e_i^T
      \end{matrix} \right )\;,
\end{eqnarray*}
where $\ind_{(\cdot)}$ is the indicator function. Using $d_i \leq \dmax$ and structure of matrices
in the summation in the last term, it can be easily verified that the $\| \cdot \|_2$ norm of the resulting
matrix is at most $1/k\dmax$.  Now we can apply the results of Lemma~\ref{lem:matrixbennet} to obtain 
a bound on $\big\|\sum_{i<j}Z^{ij}\big\|_2 = \big\|\sum_{i<j}\tZ^{ij}\big\|_2$: 
\begin{eqnarray*}
	\prob \left(\; \left \| \sum_{i<j} Z^{ij} \right \| \geq t \;\right) & \leq& 
	2 \numitems \exp \left ( \frac{-t^2/2}{(1/\resamp\dmax) + (t/\sqrt{\resamp\dmax^2})}\right )\;.
\end{eqnarray*}
\SN{Under our assumption that $\dmax \geq \log \numitems$ and 
choosing $t=C\sqrt{\log n/(k\dmax)}$, the tail probability is bounded by 
$2\numitems\exp\{-(C^2\log n/2)(1/(1+C))\}$.
Hence, we get the desired bound that 
$\|\Delta-D\|_2 \leq C\sqrt{\log \numitems/(\resamp\dmax)}$ 
with probability at least $1-2n^{-C/4+1}$, where we have used the fact
that $C \geq 8$.}

Now we are left to prove that the condition \eqref{eq:matrixbernsteincondition} holds. 
A quick calculation shows that
\begin{equation}
  \label{eq:Apower}
  (A^{ij})^p = \begin{cases}
    (A^{ij})^2 & \text{for $p$ even}\;, \\
    A^{ij} & \text{for $p$ odd}\;.
    \end{cases}
\end{equation}
Furthermore, we can verify that the eigenvalues of $A^{ij}$ are either
$1$ or $-1$. Hence, $(A^{ij})^p \preceq (A^{ij})^2$ for all $p \geq
1$. Thus, given the fact that $\tZ^{ij} = \Delta_{ij} A^{ij}$ we have
that $\E[ (\tZ^{ij})^p] = \E[ \Delta^p_{ij} (A^{ij})^p ] \preceq |\E[
\Delta_{ij}^p]| (A^{ij})^2$ for all $p$. This fact follows since for
any constant $c \in \reals$, $c A^{ij} \preceq |c| (A^{ij})^2$ and $c
(A^{ij})^2 \preceq |c| (A^{ij})^2$. Hence, coupling these observation
with the identities presented in equation~\eqref{eq:Apower} we have
\begin{eqnarray*}
      \E\big[(\tZ^{ij})^p\big] &\preceq& \E\big[|\Delta_{ij}|^p\big] (A^{ij})^2 \;, 
\end{eqnarray*}
where we used Jensen's inequality for $|\E[\Delta_{ij}^p]| \leq \E[ |\Delta_{ij}|^p]$. 

Next, it remains to construct a bound on $\E |\Delta_{ij}^p|$: 
\begin{eqnarray}
	\label{eq:boundmomentp}
	\E \big[ |\Delta_{ij}|^p \big] &\leq& \frac{p!}{2} \Big(\frac{1}{\sqrt{\resamp\dmax^2}}\Big)^p\;.
\end{eqnarray}
From \eqref{eq:f1}, we have $\Delta_{ij} = P_{ij} - \tP_{ij} = \frac{1}{k\dmax} C_{ij}$. Therefore, 
\begin{align*}
\E \big[| \Delta_{ij} |^p\big] & = (1/\resamp \dmax)^p\, \E \big[|C_{ij}|^p\big]. 
\end{align*}
Applying Azuma-Hoeffding's inequality to $C_{ij}$, we have that 
\begin{align*}
  \prob \left (\frac{1}{\resamp \dmax} |C_{ij}| \geq t \right )
  &\leq 2 \exp(-2 t^2 \dmax^2 \resamp)\;.
\end{align*}
That is, $\frac{1}{\resamp \dmax} C_{ij}$ is a sub-Gaussian random variable. And therefore, it follows 
that for $p \geq 2$, 
\begin{eqnarray*}
  \E\Big[\big|\frac{1}{\resamp \dmax} C_{ij}\big|^p\Big] &\leq& \frac{p!}{2} \Big(\frac{1}{\sqrt{\resamp\dmax^2}}\Big)^p\;.
\end{eqnarray*}
This proves the desired bound in \eqref{eq:boundmomentp}.

\subsubsection{Proof of Lemma \ref{lem:specGap}}
\label{sec:specGap}

By Lemma~\ref{lem:boundDel}, we have \SN{for some $C \geq 8$} that
\begin{eqnarray*}
  1-\rho &=& 1-\lmax(\tP)-\|\Delta\|_2\sqrt{b} \\
  &\geq& 1-\lmax(\tP)-C\sqrt{b\log n/(k\dmax)}
\end{eqnarray*}
with probability at least $1 - 4 n^{-C/8}$.
In this section we prove that there is a positive gap: 
$(\dmin/2\,b^2\,\dmax)\,\eiggap$. 
We will first prove that  
\begin{eqnarray}
	1-\lmax(\tP)&\geq& \frac{\eiggap\,\dmin}{b^2\,\dmax}\;. \label{eq:spectralgap}
\end{eqnarray}
This implies that we have the desired eigengap for 
$k \geq 4 C^2 b^5 \dmax \log n\, (1/\dmin \eiggap)^2$ such that 
$C\sqrt{b\log n/(k\dmax)} \leq (\dmin/2\,b^2\,\dmax)\,\eiggap$.

To prove \eqref{eq:spectralgap}, we use comparison theorems \citep{DS93}, 
which bound the spectral gap of the Markov chain $\tP$ of interest 
using a few comparison inequalities related to 
a more tractable Markov chain, which is the simple random walk on the graph.  
We define the transition matrix of the simple random walk on the graph $G$ as 
\begin{eqnarray*}
	\label{eq:defQ}
	Q_{ij}&=&\frac{1}{d_i}\;\;\text{   for }(i,j)\in E \;, 
\end{eqnarray*}
and the stationary distribution of this Markov chain is $\mu(i)=d_i/\sum_jd_j$.
Further, since the detailed balance equation is satisfied, $Q$ is a reversible Markov chain. 
Formally, $\mu(i)Q_{ij}=1/\sum_\ell d_\ell=\mu(j)Q_{ji}$ for all $(i,j)\in E$.  

The following key lemma is a special case of a more general result \citep{DS93}
proved for two arbitrary reversible Markov chains, which
are not necessarily defined on the same graph.  For completeness, we
provide a proof of this lemma later in this section, following a
technique similar to the one in \citep{BGPS05} used to prove a similar
result for a special case when the stationary distribution is uniform.
\begin{lemma}
    \label{lem:comparison}
    Let $Q,\mu$ and $\tP,\tpi$ be reversible Markov chains on a finite set $[n]$ 
    representing random walks on a 
    graph $G=([n],E)$, i.e.  
    $\tP(i,j)=0$ and $Q(i,j)=0$ if $(i,j)\notin E$.
    For $\alpha\equiv\min_{(i,j)\in E}\{\tpi(i)\tP_{ij}/\mu(i)Q_{ij}\}$ and 
    $\beta\equiv\max_i\{\tpi(i)/\mu(i)\}$,   
    \begin{eqnarray}
	\frac{1-\lmax(\tP)}{1-\lmax(Q)}&\geq&\frac{\alpha}{\beta}\;.\label{eq:comparison}
    \end{eqnarray}
\end{lemma}
By assumption, we have $\eiggap\equiv 1-\lmax(Q)$. 
To prove that there is a positive spectral gap for 
the random walk of interest as in \eqref{eq:spectralgap}, 
we are left to bound $\alpha$ and $\beta$. 
We have $\mu(i)Q_{ij}=1/\sum_\ell d_\ell\leq1/|E|$ and 
$\mu(i)\geq(d_i/|E|)$. 
Also, by assumption that $\max_{i,j} w_i/w_j \leq b$, 
we have $\tpi(i)\tP_{ij}=w_iw_j/(\dmax(w_i+w_j)\sum_\ell w_\ell)\geq1/(bn\dmax)$ 
and $\tpi(i)=w_i/\sum_\ell w_\ell\leq b/n$.  
Then, $\alpha=\min_{(i,j)\in E}\{\tpi(i)\tP_{ij}/\mu(i)Q_{ij}\}\geq |E|/(nb\dmax)$ 
and $\beta=\max_i\{\tpi(i)/\mu(i)\}\leq b|E|/n\dmin$. 
Hence, $\alpha/\beta \geq \dmin/(\dmax b^2)$ and this finishes the proof of the bound in \eqref{eq:spectralgap}. 

\subsubsection{Proof of Lemma \ref{lem:comparison}}
\label{sec:proofcomparison}
Since $1-\lmax = \min\{1-\lambda_2,1+\lambda_n\}$, 
we will first show that 
$1-\lambda_2(Q)\leq(\beta/\alpha)(1-\lambda_2(\tP))$ and 
$1+\lambda_n(Q)\leq(\beta/\alpha)(1+\lambda_n(\tP))$. 
The desired bound in \eqref{eq:comparison} immediately follows from the fact that 
$\min\{a,b\}\leq\min\{a',b'\}$ if $a\leq b$ and $a'\leq b'$. 

A reversible Markov chain $Q$ is self-adjoint in $L_2(\mu)$.
Then, the second largest eigenvalue $\lambda_2(Q)$ can be 
represented by the Dirichlet form $\cE$ defined as 
\begin{eqnarray*}
    \cE^Q(\phi,\phi)\;\;\equiv\;\;\big\<(I-Q)\phi,\phi\big\>_\mu\;\;=\;\;\frac12\sum_{i,j}(\phi(i)-\phi(j))^2\mu(i)Q(i,j) \;.
\end{eqnarray*}
For $\lambda_n(Q)$, we use 
\begin{eqnarray*}
    \cF^Q(\phi,\phi)\;\;\equiv\;\;\big\<(I+Q)\phi,\phi\big\>_\mu\;\;=\;\;\frac12\sum_{i,j}(\phi(i)+\phi(j))^2\mu(i)Q(i,j) \;.
\end{eqnarray*}
Following the usual variational characterization of the eigenvalues (see, for instance, \citep{HJ85}, p. 176) gives 
\begin{eqnarray}
    1-\lambda_2(Q)&=&\min_{\phi\perp\ones} \frac{\cE^Q(\phi,\phi)}{\<\phi,\phi\>_\mu}\;,\label{eq:variational1}\\
    1+\lambda_n(Q)&=&\min_{\phi} \frac{\cF^Q(\phi,\phi)}{\<\phi,\phi\>_\mu}\;.\label{eq:variational2}
\end{eqnarray}

By the definitions of $\alpha$ and $\beta$, we have $\tpi(i,)\tP(i,j)\geq\alpha\mu(i)Q(i,j)$ 
and $\tpi(i)\leq\beta\mu(i)$ for all $i$ and $j$, which implies
\begin{eqnarray*}
    \cE^\tP(\phi,\phi)&\geq&\alpha\cE^Q(\phi,\phi)\;,\\
    \cF^\tP(\phi,\phi)&\geq&\alpha\cF^Q(\phi,\phi)\;,\\
    \<\phi,\phi\>_\tpi&\leq&\beta\<\phi,\phi\>_\mu\;.
\end{eqnarray*}
Together with \eqref{eq:variational1}, 
this implies $1-\lambda_2(Q)\leq(\beta/\alpha)(1-\lambda_2(\tP))$ and 
$1+\lambda_n(Q)\leq(\beta/\alpha)(1+\lambda_n(\tP))$. 
This finishes the proof of the desired bound.

\subsection{Proof of Theorem \ref{thm:mainER}: Random sampling}
\label{sec:mainERproof}

Given the proof of Theorem~\ref{thm:maingeneral} in the previous section, 
we only need to prove that for an Erd\"os-Renyi graph with average degree 
$d\geq C'\log n$ the following are true: 
\begin{eqnarray}
	(1/2)d&\leq&d_i\;\;\leq\;\;(3/2)d \;, \label{eq:ERdegreebound}\\
	1/2&\leq&\eiggap\;. \label{eq:EReigbound}
\end{eqnarray}
Then, it follows that $\degratio\leq 3$ and $(1/2)d\leq \dmin \leq
\dmax\leq (3/2)d$.  By Theorem~\ref{thm:maingeneral}, it follows that
\begin{eqnarray*}
	\frac{\|\pi-\tpi\|}{\|\tpi\|} &\leq& 6 C b^{5/2}\sqrt{\frac{\log n}{k\,d}}\;,
\end{eqnarray*}
with probability at least $1-4 n^{-C/8}$ for some positive constant $C
\geq 8$ and for $k d \geq 288 C^2 b^5 \log n$

We can apply standard concentration inequalities to establish equation
\eqref{eq:ERdegreebound}.  Apply Chernoff's inequality, we get
$\prob\big(\, |d_i-d|>(1/2)d\,\big) \leq 2e^{-d/16}$.  Hence, for
$d\geq C'\log n$, equation \eqref{eq:ERdegreebound} is true with
probability at least $1-2n^{-C'/16}$.

Finally, we finish the proof with a result on the lower bound of the spectral gap
$\eiggap=1-\lmax(D^{-1}B)$.
\begin{lemma}
  \label{lem:specgap}
  Consider a random graph $G$ drawn from the Er\"dos-Renyi distribution $G(n,d/n)$.
  Then if $d \geq 10 C^2 \log n$, we have
  $\eiggap \geq 1/2$
  with probability at least $1-n^{-C n/(n-d)/8}$
\end{lemma}
The proof of this result can be found in Appendix~\ref{app:specgap}.

\subsection{Proof of Theorem \ref{thm:lowerbound}: Information-theoretic lower bound}
\label{sec:prooflowerbound}

In this section, we prove Theorem~\ref{thm:lowerbound} 
using an information-theoretic method that allows us to reduce the 
stochastic inference problem into a multi-way hypothesis testing problem. 

This estimation problem can be reduced to the following hypothesis testing problem. 
Consider a set $\{{\tpi}^{(1)},\ldots,{\tpi}^{(\npack(\distpack))}\}$ 
of $\npack(\distpack)$ vectors on the standard orthogonal simplex 
which are separated by $\distpack$, such that 
$\|{\tpi}^{(\ell_1)}-{\tpi}^{(\ell_2)}\|\geq\distpack$ for all $\ell_1\neq\ell_2$. 
To simplify the notations, we are going to use $\npack$ 
as a shorthand for $\npack(\distpack)$.
Suppose we choose an index $\idxpack\in\{1,\ldots,M\}$ uniformly at random. 
Then, we are given noisy outcomes of pair-wise comparisons 
with $w={\tpi}^{(\idxpack)}$ 
from the BTL model. 
We use $\obs$ to denote this set of observations. 
Let $\pi$ be the estimation produced by an algorithm using the noisy observations. 
Given this, the best estimation of the ``index'' is $\idxest$, where 
$\idxest=\arg\min_{\ell\in[\npack]}{\|\pi-{\tpi}^{(\ell)}\|}$.

By construction of our packing set, when 
we make a mistake in the hypothesis testing, our estimate is at least 
$\distpack/2$ away from the true weight ${\tpi}^{(\idxpack)}$. 
Precisely, $\idxest\neq\idxpack$ implies that $\|\pi-{\tpi}^{(\idxpack)}\|\geq\distpack/2$. 
Then, 
\begin{eqnarray}
	\E\big[\,\|\pi-{\tpi}^{(\idxpack)}\|\,\big] &\geq& \frac{\distpack}{2}\prob\big(\idxest\neq\idxpack\big)\nonumber\\
	&\geq& \frac{\distpack}{2} \Big\{1-\frac{I(\idxest;\idxpack)+\log 2}{\log \npack}\Big\}\;,\label{eq:fano}
\end{eqnarray}
where $I(\cdot;\cdot)$ denotes the mutual information between two random variables 
and the second inequality follows from Fano's inequality. 

These random vectors form a Markov chain 
$\idxpack\,$---$\,{\tpi}^{(\idxpack)}\,$---$\,\obs\,$---$\,\pi\,$---$\,\idxest\,$, 
where $X\,$---$\,Y\,$---$\,Z$ indicates that $X$ and $Z$ 
are conditionally independent given $Y$. 
Let $\prob_{\idxpack,\obs}(\ell,x)$ denote the joint probability function, 
and  $\prob_{\obs|\idxpack}(x|\ell)$, $\prob_{\idxpack}(\ell)$ and $\prob_{\obs}(x)$ 
denote the conditional and marginal probability functions.
Then, by data processing inequality for a Markov chain, we get  
\begin{eqnarray}
	I(\idxpack;\idxest) &\leq& I(\idxpack ;\obs)\nonumber\\
	&=& \E_{\idxpack,\obs} \Big[\log\Big(\frac{\prob_{\idxpack,\obs}(\idxpack,\obs)}{\prob_{\idxpack}(\idxpack)\prob_{\obs}(\obs)}\Big)\Big] \nonumber\\
	&=& \frac{1}{\npack}\sum_{\ell\in[\npack]} \E_{\obs} \Big[\log\Big(\frac{\prob_{\obs|\idxpack}(\obs|\ell)}{\prob_{\obs}(\obs)}\Big)\Big] \nonumber\\
	&=& \frac{1}{\npack}\sum_{\ell\in[\npack]} \E_{\obs} \Big[\log\Big(\frac{\prob_{\obs|\idxpack}(\obs|\ell)}{\sum_{\ell_2\in[\npack]}\prob_{\obs|\idxpack}(\obs|\ell_2)\prob(\ell_2)}\Big)\Big] \nonumber\\
	&\leq& \frac{1}{\npack}\sum_{\ell\in[\npack]} \sum_{\ell_2\in[\npack]} \prob(\ell_2) \E_{\obs} \Big[\log\Big(\frac{\prob_{\obs|\idxpack}(\obs|\ell)}{\prob_{\obs|\idxpack}(\obs|\ell_2)}\Big)\Big] \nonumber\\
	&=& \frac{1}{\npack^2}\sum_{\ell_1,\ell_2}\KL\Big(\,\prob_{\obs|\idxpack}(\obs|\ell_1)\,\Big\|\,\prob_{\obs|\idxpack}(\obs|\ell_2)\,\Big)\;,\label{eq:mutualinformation}
\end{eqnarray}
where $\KL(\cdot\|\cdot)$ is the Kullback-Leibler (KL) divergence  
and the inequality follows from the concavity of logarithm and 
Jensen's inequality.

The KL divergence between the observations coming from two different BTL models 
depend on how we sample the comparisons. 
We are sampling each pair of items for comparison with probability 
$\numcomp/\numitems$, and we are comparing each of these sampled pairs 
$k$ times. 
Let $\obs_{ij}$ denote the outcome of $k$ comparisons for a sampled 
pair of items $(i,j)$. 
To simplify notations, we drop the subscript $\obs|\idxpack$ whenever it is clear from the context. 
Then, 
\begin{eqnarray}
	\KL\big(\,\prob(\obs|\ell_1)\,\big\|\,\prob(\obs|\ell_2)\,\big) &=& 
	\frac{\numcomp}{\numitems}\sum_{1\leq i<j\leq \numitems} \KL\big(\prob(\obs_{ij}|\ell_1)\,\big\|\,\prob(\obs_{ij}|\ell_2)\big) \nonumber\\
	&\leq& 2\,\numitems^2\,k\,\numcomp\, \big\| ({\tpi}^{(\ell_1)}-{\tpi}^{(\ell_2)}\big\|^2\label{eq:KL}\;,
\end{eqnarray}
where in the last inequality we used the fact that 
\begin{eqnarray*}
	\KL\big(\,\prob(X_{ij}|\ell_1)\,\|\,\prob(X_{ij}|\ell_2)\,\big) &\leq& 
	\frac{k\big({\tpi}^{(\ell_2)}_j({\tpi}^{(\ell_1)}_i-{\tpi}^{(\ell_2)}_i)^2+
	{\tpi}^{(\ell_2)}_i({\tpi}^{(\ell_1)}_j-{\tpi}^{(\ell_2)}_j)^2\big)} 
	{{\tpi}^{(\ell_2)}_i{\tpi}^{(\ell_2)}_j\big({\tpi}^{(\ell_1)}_i+{\tpi}^{(\ell_1)}_j\big)} \\
	&\leq&2k\numitems^2 \Big(({\tpi}^{(\ell_1)}_i-{\tpi}^{(\ell_2)}_i)^2 + ({\tpi}^{(\ell_1)}_j-{\tpi}^{(\ell_2)}_j)^2\Big)\;,
\end{eqnarray*}
for $k$ independent trials of 
Bernoulli random variables, and 
${\tpi}^{(\ell)}_i\geq1/(2\numitems)$ for all $i$ and $\ell$ which follows 
from our construction of the packing set in Lemma~\ref{lem:packing} and our choice of $\distpack$.  

The remainder of the proof relies on the following key technical lemma, 
on the construction of a suitable packing set 
that has enough number of entries which are reasonably separated. 
This is proved in Section~\ref{sec:proofpacking}. 
\begin{lemma}
	\label{lem:packing}
	For $\numitems\geq90$ and for any positive $\distpack\leq1/2\sqrt{10\numitems}$, there exists a set of 
	$\numitems$-dimensional vectors $\{{\tpi}^{(1)},\ldots,{\tpi}^{(\npack)}\}$ 
	with cardinality $\npack=e^{\numitems/128}$ such that $\sum_i{\tpi}^{(\ell)}_i=1$ and 
	\begin{eqnarray*}
		\frac{1-2\distpack\sqrt{10\numitems}}{\numitems} &\leq& {\tpi}^{(\ell)}_i \;\;\leq\;\; \frac{1+2\distpack\sqrt{10\numitems}}{\numitems}\;, 
	\end{eqnarray*}
	for all $i\in[\numitems]$ and $\ell\in[\npack]$, and 
	\begin{eqnarray*}
		\distpack &\leq& \|{\tpi}^{(\ell_1)}-{\tpi}^{(\ell_2)}\| \;\;\leq\;\; \sqrt{13}\distpack\;,
	\end{eqnarray*}
	for all $\ell_1\neq\ell_2$. 
\end{lemma}
Substituting this bound in Eqs.~\eqref{eq:KL}, \eqref{eq:mutualinformation}, and \eqref{eq:fano}, 
we get 
\begin{eqnarray*}
	\max_{\ell\in[\npack]}\E[\,\|\pi-{\tpi}^{(\ell)}\|\,] &\geq& \E[\,\|\pi-{\tpi}^{(\idxpack)}\|\,] \\
	&\geq& \frac{\distpack}{2}\Big\{1-\frac{3328 \numitems^2k\numcomp\distpack^2+128\log 2}{\numitems}\Big\}\;.
\end{eqnarray*}
Choosing $\distpack=(\upbound-1)/(30\sqrt{10}(\upbound+1)\sqrt{k \numcomp \numitems})$, 
we know that $3328 \numitems^2k\numcomp\distpack^2+128\log 2\leq (1/2)\numitems$ 
for all $b$ and all $\numitems\geq682$. 
This implies that 
\begin{eqnarray*}
	\max_{\ell\in[\npack]}\E\big[\,\|\pi-{\tpi}^{(\ell)}\|\,\big] &\geq& \frac{(\upbound-1)}{120(\upbound+1)\sqrt{10k\numcomp\numitems}}\;.
\end{eqnarray*}
From Lemma~\ref{lem:packing}, it follows that 
$\|{\tpi}^{(\ell)}\|\leq2/\sqrt{\numitems}$ for all $\ell$. 
Then, scaling the bound by $1/\|{\tpi}^{(\ell)}\|$, 
the normalized minimax rate is lower bounded by $(\upbound-1)/(240(\upbound+1)\sqrt{10k\numcomp})$.  
Also, for this choice of $\distpack$, the dynamic range is at most $\upbound$. 
From Lemma~\ref{lem:packing}, the dynamic range is upper bounded by
\begin{eqnarray*}
	\max_{\ell,i,j}\frac{{\tpi}^{(\ell)}_i}{{\tpi}^{(\ell)}_j} &\leq& \frac{1+2\distpack\sqrt{10n}}{1-2\distpack\sqrt{10n}}\;.
\end{eqnarray*}
This is monotonically increasing in $\distpack$ for $\distpack<1/(2\sqrt{10\numitems})$. 
Hence, for $\distpack\leq(\upbound-1)/((\upbound+1)2\sqrt{10\numitems})$,
which is always true for our choice of $\distpack$, 
the dynamic range is upper bounded by $\upbound$.
This finishes the proof of the desired bound on normalized minimax error rate for general $\upbound$. 


\subsubsection{Proof of Lemma~\ref{lem:packing}}
\label{sec:proofpacking}

We show that a random construction succeeds in generating a set 
of $\npack$ vectors on the standard orthogonal simplex satisfying the conditions 
with a strictly positive probability. 
Let $\npack=e^{\numitems/128}$ and for each $\ell\in[\npack]$, we 
construct independent random vectors $\tpi^{(\ell)}$ according to the following procedure. 
For a positive $\alpha$ to be specified later, 
we first draw $\numitems$ random variables uniformly from 
$[(1-\alpha\distpack\sqrt{\numitems})/\numitems,(1+\alpha\distpack\sqrt{\numitems}))/\numitems]$. 
Let $Y^{(\ell)}=[Y^{(\ell)}_1,\ldots,Y^{(\ell)}_{\numitems}]$ 
denote this random vector in $\numitems$ dimensions. 
Then we project this onto the $\numitems$-dimensional simplex by setting 
\begin{eqnarray*}
	\tpi^{(\ell)} &=& Y^{(\ell)}+(1/n-\bY^{(\ell)})\ones \;,
\end{eqnarray*}
where $\bY^{(\ell)}=(1/\numitems)\sum_{i}Y^{(\ell)}_i$. 
By construction, the resulting vector is on the standard orthogonal simplex: $\sum_i\tpi^{(\ell)}_i=1$. 
Also, applying Hoeffding's inequality for $\bY^{(\ell)}$, we get that 
\begin{eqnarray*}
	\prob\Big(\Big|\bY^{(\ell)}-\frac{1}{\numitems}\Big|>\frac{\alpha\distpack}{\sqrt{\numitems}}\Big)&\leq&2e^{-\numitems/2}\;. 
\end{eqnarray*}
By union bound, this holds uniformly for all $\ell$ with probability at least 
$1-2e^{-63\numitems/128}$. 
In particular, this implies that 
\begin{eqnarray}
	\frac{1-2\alpha\distpack\sqrt{\numitems}}{\numitems}&\leq&\tpi^{(\ell)}_i\;\;\leq\;\;\frac{1+2\alpha\distpack\sqrt{\numitems}}{\numitems}\;,
	\label{eq:entrybound}
\end{eqnarray}
for all $i\in[\numitems]$ and $\ell\in[\npack]$.

Next, we use standard concentration results to bound the distance between two vectors:
\begin{eqnarray*}
	\big\|\tpi^{(\ell_1)}-\tpi^{(\ell_2)}\big\|^2 &=& \big\|Y^{(\ell_1)}-Y^{(\ell_2)}\big\|^2 -\numitems(\bY^{(\ell_1)}-\bY^{(\ell_2)})^2 
\end{eqnarray*}
Applying Hoeffding's inequality for the first term, 
we get $\prob\big(|\sum_i(Y^{(\ell_1)}_i-Y^{(\ell_2)}_i)^2-(2/3)\alpha^2\distpack^2|\geq(1/2)\alpha^2\distpack^2\big) \leq 2e^{-\numitems/32}$. 
Similarly for the second term, we can show that 
$\prob\big(|\sum_i(Y^{(\ell_1)}_i-Y^{(\ell_2)}_i)|\geq(1/4)\alpha\distpack\sqrt{\numitems}\big) \leq 2e^{-\numitems/32}$. 
Substituting these bounds, we get  
\begin{eqnarray}
	\frac{1}{10}\alpha^2\distpack^2&\leq&\|\tpi^{(\ell_1)}-\tpi^{(\ell_2)}\|^2\;\;\leq\;\;\frac{13}{10}\alpha^2\distpack^2\;,
	\label{eq:normbound}
\end{eqnarray}
with probability at least $1-4e^{-\numitems/32}$. 
Applying union bound over ${{\npack}\choose{2}}\leq e^{\numitems/64}$ pairs of vectors, 
we get that the lower and upper bound holds for all pairs $\ell_1\neq\ell_2$ 
with probability at least $1-4e^{-\numitems/64}$. 

The probability that both conditions \eqref{eq:entrybound} and 
\eqref{eq:normbound} are satisfied is at least 
$1-4e^{-\numitems/64}-2e^{-63\numitems/128}$.
For $\numitems\geq90$, the probability of success is strictly positive. 
Hence, we know that there exists at least one set of vectors that satisfy the conditions. 
Setting $\alpha=\sqrt{10}$, we have constructed a set that satisfy all the conditions.

\subsection{Proof of Theorem~\ref{thm:mle}: Finite sample analysis of MLE}
The proof of this theorem will follow in two parts. First we will show that
if the gradient of the loss $\nabla \Loss$ evaluated at $\thetaparam$ is small, then the error between $\thetaparam$ and $\thetahat$ is also small. To that
end we begin with a simple inequality:
\begin{equation*}
  \Loss(\thetahat) \leq \Loss(\thetastar).
\end{equation*}
Let $\Delta = \thetahat - \thetastar$. We can add and subtract $\inprod{\gradloss(\thetastar)}{\Delta}$ from the above equation to obtain
\begin{align*}
  \Loss(\thetastar + \Delta) - \Loss(\thetastar) - \inprod{\gradloss(\thetastar)}{\Delta} & \leq \inprod{\gradloss(\thetastar)}{\Delta}.
\end{align*}
\emph{Now assume $\|\gradloss(\thetastar)\|_2 \leq c$.} By the Cauchy-Schwartz inequality we have that
\begin{align*}
  \Loss(\thetastar + \Delta) - \Loss(\thetastar) - \inprod{\gradloss(\thetastar)}{\Delta} & \leq c \|\Delta\|_2.
\end{align*}
Therefore, we if we prove that
\begin{align}
  \label{eq:sc}
  \Loss(\thetastar + \Delta) - \Loss(\thetastar) - \inprod{\gradloss(\thetastar)}{\Delta} & \geq \frac{\mu}{2} \|\Delta\|_2^2,
\end{align}
then we immediately have that $\|\Delta\|_2 \leq 2 c/\mu$. We now proceed
to establish the above inequality.
\subsubsection{Proof of Equation~\ref{eq:sc}}
\label{subsec:proofofsc}
By Taylor's theorem and the definition of $\Loss$ from equation~\ref{defn:loss}
for some $v \in [0,1]$ we have
\begin{align*}
  \Loss(\thetastar + \Delta) - \Loss(\thetastar) -
  \inprod{\gradloss(\thetastar)}{\Delta} & = \frac{1}{2 \numobs}
  \sum_{l=1}^\numobs \frac{\exp(\myexp)}{(1+\exp(\myexp))^2} (
  \inprod{\Delta}{\design_l} )^2.
\end{align*}
Now, by assumption $\sum_i \thetaparam_i = \sum_i \thetahat_i = 0$; and
$\thetaparam_{\max} - \thetaparam_{\min}$ and $\thetahat_{\max} - \thetahat_{\min} \leq \log(b)$  so that $|\myexp| \leq \log(b)$. Therefore,
\begin{align*}
  \Loss(\thetastar + \Delta) - \Loss(\thetastar) -
  \inprod{\gradloss(\thetastar)}{\Delta} & \geq \frac{1}{2 \numobs}
  \sum_{l=1}^\numobs \frac{b}{(1+b)^2} (  \inprod{\Delta}{\design_l} )^2.
\end{align*}
Thus, what remains is to establish a lower-bound on
\begin{align*}
  \frac{1}{\numobs} \sum_{l=1}^\numobs (  \inprod{\Delta}{\design_l} )^2.
\end{align*}
We appeal to the following lemma for the lower-bound.
\begin{lemma}
  \label{lem:rsc}
  Given $\numobs > 12 \numitems \log \numitems$ i.i.d. samples
  $y_l,x_l$ we have that
  \begin{align*}
    \frac{1}{\numobs} \sum_{l=1}^\numobs (  \inprod{\Delta}{\design_l} )^2 &
    \geq \frac{1}{3 \numitems} \| \Delta \|_2^2
  \end{align*}
  with probability at least $1 - 1/\numitems$.
\end{lemma}
Finally, we present the following lemma that establishes an upper-bound on $\| \gradloss(\thetaparam) \|_2$.
\begin{lemma}
  \label{lem:gradbound}
  Given $\numobs$ observations $(\mleobs_l,\design_l)$ we have that
  \begin{equation*}
    \| \gradloss(\thetaparam) \|_2 \leq 2 \sqrt{\frac{\log \numitems}{\numobs}}
  \end{equation*}
  with probability at least $1 - 1/\numitems$.
\end{lemma}
Therefore, putting everything together we have that
\begin{align*}
  \| \Delta \|_2 & \leq 6 (1+b)^2/b \sqrt{\frac{\numitems^2 \log \numitems}{\numobs}},
\end{align*}
which establishes the desired result.
\subsubsection{Proof of Lemma~\ref{lem:rsc}}
To prove this lemma we note that
\begin{equation*}
  \frac{1}{\numobs} \sum_{l=1}^\numobs (  \inprod{\Delta}{\design_l} )^2 = 
  \frac{1}{\numobs} \sum_{l=1}^\numobs \Delta^T \design_l \design_l^T \Delta.
\end{equation*}
Thus, it is sufficient to prove a lower-bound on $\lambda_{\min}(\frac{1}{\numobs} \sum_{l=1}^\numobs \design_l \design_l^T)$.
In order to do so we may again appeal
to recent results on random matrix theory~\cite{Jo11}.
\begin{lemma}[Theorem 1.4~\citep{Jo11}]
  Consider a finite sequence $\{X_k\}$ of independent, random, self-adjoint matrices with dimensions $d$. Assume that each random matrix satisfies
$\E X_k = 0$ and $\lambda_{\max}(X_k) \leq R$ almost surely. Then, for all $t \geq 0$,
\begin{align}
  \label{eq:matrixbernstein}
  \prob \left \{ \lambda_{\max} \left ( \sum_k X_k \right ) \geq t \right \} & \leq d \cdot \exp \left ( \frac{-t^2/2}{\sigma^2 + R t/3} \right ) \; \text{ where } \; \sigma^2 \defn \| \sum_k \E(X_k^2) \|,
\end{align}
and $\| X \|$ for a matrix $X$ represents the operator norm of $X$ or
its larges singular value.
\end{lemma}
In order to apply the above lemma we let $X_l = x_l x_l^T - 2/\numitems (I - \ones \ones^T/\numitems)$. Therefore, the $X_l$ are zero-mean, i.i.d., and symmetric. Furthermore, $\|X_l\| \leq 2$ and $\E X_l^2 = 4/\numitems (I - \ones \ones^T/\numitems) - 4/\numitems^2 (I- \ones \ones^T/\numitems)$. Therefore, applying the above lemma to both $X_l$ and $-X_l$ yields the inequality
\begin{equation*}
  \prob \left \{ \|\sum_l X_l/\numobs\| \geq t \right \} \; \leq 2 \numitems \exp \left ( \frac{-t^2/2}{\frac{4}{\numitems \numobs} + 2 t/(3 \numobs)}\right ).
\end{equation*}
Thus, with probability at least $1 - 1/\numitems$,
\begin{align*}
  \|\frac{1}{\numobs} \sum_l X_l\| & \leq \max( 4 \sqrt{\frac{2 \log \numitems}{\numitems \numobs}},8/3 \frac{\log \numitems}{\numobs} ).
\end{align*}
Hence, as long as $12 \numitems \log \numitems < \numobs$, then
\begin{align*}
  \|\frac{1}{\numobs} \sum_l X_l\| & \leq 4 \sqrt{\frac{2 \log \numitems}{\numitems \numobs}},
\end{align*}
with probability at least $1 - 1/\numitems$.

With the above result in hand we now have that
\begin{equation*}
  \| \frac{1}{\numobs} \sum_{l=1}^\numobs \design_l \design_l^T - \frac{2}{\numitems} (I - \ones \ones^T/\numitems) \| \leq 4 \sqrt{\frac{2 \log \numitems}{\numitems \numobs}}.
\end{equation*}
Therefore,
\begin{equation*}
  \frac{1}{\numobs} \sum_{l=1}^\numobs \Delta^T \design_l \design_l^T \Delta \geq \frac{2}{\numitems} \|\Delta\|_2^2(1 - 2 \sqrt{\frac{2 \numitems \log \numitems}{\numobs}}),
\end{equation*}
where we have used the fact that $\Delta = \thetahat - \thetaparam$
and $\sum_i \thetahat_i = \sum_i \thetaparam_i = 0$. Recalling that,
$\numobs > 12 \numitems \log \numitems$ the above inequality can be
lower bounded by
$\frac{1}{3 \numitems} \| \Delta\|_2^2$,
establishing the desired result.
\subsubsection{Proof of Lemma~\ref{lem:gradbound}}
To establish this result we will proceed by showing each individual
element of $\gradloss$ is upper bounded by $2 \sqrt{\log \numitems
  /(\numitems \numobs)}$ with high probability. Recall that
\begin{equation*}
  \gradloss = \frac{1}{\numobs} \sum_{l=1}^\numobs x_l (\E [\obs_l | x_l] - \obs_l).
\end{equation*}
Consequently, focusing on a single component ${\gradloss}_k$ we have that
\begin{equation*}
  {\gradloss}_k = \frac{1}{\numobs} \sum_{l=1}^\numobs (x_l)_k (\E [\obs_l | x_l] - \obs_l).
\end{equation*}
Thus, the $k^{th}$ component of $\gradloss$ is the average over
$\numobs$ independent mean zero random variables that are
upper-bounded by $1$ and that each have variance upper-bounded by
$1/\numitems$. Therefore, an application of Bernstein's inequality yields
\begin{align*}
  \prob( | {\gradloss}_k | \geq t ) & \leq 2 \exp \left (    \frac{-t^2}{\frac{2}{\numitems \numobs} + \frac{2t }{3 \numobs}}    \right ).
\end{align*}
Therefore,
\begin{align*}
  \prob( \| \gradloss \|_\infty \geq t ) & \leq \numitems \prob (| {\gradloss}_k | \geq t) \\
  & \leq 2 \numitems \exp \left (    \frac{-t^2}{\frac{2}{\numitems \numobs} + \frac{2t }{3 \numobs}}    \right ).
\end{align*}
Using arguments similar to those to establish the results in Section~\ref{subsec:proofofsc} we have that with probability at least $1 - 2/\numitems$
\begin{align*}
  \| \gradloss \|_\infty & \leq 2 \sqrt{\frac{\log \numitems}{\numitems \numobs}},
\end{align*}
as desired.

\section{Discussion}
\label{SecDiscuss}

The main contribution of this paper is the design and analysis of \RC: an iterative algorithm for rank aggregation using pair-wise comparisons. We established the efficacy of the algorithm by analyzing its performance when data is generated as per the popular Bradley-Terry-Luce (BTL) or Multinomial Logit (MNL) model.  We have obtained an analytic bound on the finite sample error rates between the scores assumed by the BTL model and those estimated by our algorithm.  As shown, these lead to near-optimal dependence on the number of samples required to learn the scores well by our algorithm under random selection of pairs for comparison. More generally, the comparison graph structure plays a crucial role in the performance of the algorithm.

   For a tighter analysis of the optimality of Rank Centrality, we provide numerical experiments under the BTL model and compare it to the Cramer Rao lower-bound. Comparisons with the Cramer-Rao bound in Figure \ref{fig:crb} suggests that the error achieved by Rank Centrality is indistinguishable from the fundamental Cramer-Rao lower bound, and thus suggesting it's stronger optimality properties compared to what we can establish.

\finalcomment{ For completeness, we further provided an analysis of the
  error achieved by the MLE. Building upon our 
  analysis, \cite{HOX14} shows that MLE is near order-optimal, just like
  \RC. } It is worth noting, however, that empirically the computational cost of
  \RC~seems much better than that of finding the MLE.

\newpage

%
\begin{APPENDIX}{Proof of  Lemma~\ref{lem:metric}}

\section{Proof of  Lemma~\ref{lem:metric}}
\label{sec:proofmetric}

Without loss of generality, let us consider two items $i$ and $j$ such
that $w_i>w_j$.  When we estimate a higher score for item $j$ then we
make a mistake in the ranking of these two items.  When this happens,
such that $\pi_j-\pi_i>0$, it naturally follows that $w_i-w_j \leq
w_i-w_j + \pi_j-\pi_i \leq |w_i-\pi_i|+|\pi_j-w_j|$.  For a general
pair $i$ and $j$, we have $(w_i-w_j)(\sigma_i-\sigma_j)>0$ implies
that $|w_i-w_j| \leq |w_i-\pi_i|+|w_j-\pi_j|$.  Substituting this into
the definition of the weighted distance $\dist_w(\cdot)$, and using
the fact that $(a+b)^2\leq2a^2+2b^2$, we get
\begin{eqnarray*}
      \dist_w(\sigma) &=& 
      \Big\{\frac{1}{2\numitems\|w\|^2}\sum_{i<j} (w_i-w_j)^2 \,\ind\big((w_i-w_j)(\sigma_i-\sigma_j)>0\big)\Big\}^{1/2}\\
      &\leq& \Big\{\frac{1}{\numitems\|w\|^2}\sum_{i<j} \big\{(w_i-\pi_i)^2 + (w_j-\pi_j)^2\big\}\Big\}^{1/2}\\
      &\leq& \frac{1}{\|w\|} \left \{\sum_{i=1}^{\numitems} (w_i-\pi_i)^2 \right \}^{1/2}\;.
\end{eqnarray*}
This proves that the distance $\dist_w(\sigma)$ is upper bounded by 
the normalized Euclidean distance $\|w-\pi\|/\|w\|$.

\begin{lemma}\label{extra-lem:1}
For any $0 \leq \theta \leq \ln 4/3$, 
\begin{align}
\E[\exp(\theta |C_{ij}|)] & \leq \snncomment{2} \exp\big(2kp_{ij}\theta^2/3\big). \label{eq:rc.3}
\end{align}
\end{lemma}
\noindent{\em Proof.}
\snncomment{ Note that $C_{ij}$ is zero-mean shifted binomial random
  variable $B(k_i,p_{ij})$. Therefore, by Hoeffding's bound and the
  fact that $C_{ij}$ is the sum of $k$ terms where each term is
  upper-bounded by $\max(p_{ij},1-p_{ij})$ and lower-bounded by
  $\min(p_{ij},1-p_{ij})$
  \begin{align*}
    \E [ \exp(\theta |C_{ij}|) ] & = \E [ \exp(\theta C_{ij}) \ind{C_{ij} \geq 0}] + \E [ \exp(-\theta C_{ij}) \ind{C_{ij} < 0}] \\
    & \leq \E [ \exp(\theta C_{ij}) ] +  \E [ \exp(-\theta C_{ij}) ] \\
    & = \E [ \exp(\theta C_{ij} ) ] +  \E [ \exp(-\theta C_{ij}) ] \\
    & \leq 2 \exp(\theta^2 k/8)
  \end{align*}
}
Observe that for any $x \in \mathbb R$ and $\theta > 0$, 
\begin{align*}
\exp(\theta |x|) & \leq \exp(\theta x) + \exp(-\theta x). 
\end{align*}
From this, it follows that 
\begin{align}\label{eq:rc1a}
\E[\exp(\theta |C_{ij}|)] & \leq \E[\exp(\theta C_{ij})] + \E[\exp(-\theta C_{ij})]. 
\end{align}
Now for any $\theta \in \mathbb R$, using the fact that $X_{ij}$ is Binomial distribution and $1 + x \leq \exp(x)$ for any $x \in \mathbb R$, we 
have
\begin{align}
\E[\exp(\theta C_{ij})] & = \exp(-\theta k p_{ij}) \big(1 + p_{ij} (\exp(\theta)-1)\big)^k \nonumber \\
      & \leq \exp(-\theta k p_{ij}) \exp\big(kp_{ij} (\exp(\theta)-1)\big). \label{eq:rc.2}
\end{align}
Using second-order Taylor's expansion, for any $\theta \in [-\ln 4/3, \ln 4/3]$, we obtain that 
\begin{align}
| \exp(\theta) - 1 - \theta | & \leq \frac{2}{3} \theta^2. 
\end{align}
Using above display in \eqref{eq:rc.2}, we can obtain the claimed result.

\section{Proof of Lemma~\ref{lem:specgap}}
\label{app:specgap}

Since we are interested in the eigenvalues
of $L=D^{-1}B$, we define a more tractable matrix with the same set of
eigenvalues: $\tL=D^{-1/2}BD^{-1/2}$.  Because $\tL$ is a symmetric
matrix, the eigenvalues are the same as the singular values up to a
sign.  Let $\sigma_1(\tL)\geq\sigma_2(\tL)\geq\ldots$ denote the
ordered singular values of $\tL$.
 \SN{Note that the matrix $D^{-1/2} B D^{-1/2}$ has largest singular
   value equal to $1$. Therefore,}
 \begin{equation*}
    \sigma_2(\tL) \; \leq \|D^{-1/2} B D^{-1/2} - 1 1^T/n\|_2
  \end{equation*}
  because the vector $1/\sqrt{n}$ has unit norm. Decomposing the above we have that
  \begin{align*}
    \|D^{-1/2} B D^{-1/2} - 1 1^T/n\|_2 & \leq \|B/d - 1 1^T/n\|_2 + \|B/d - D^{-1/2} B D^{-1/2} \|_2
  \end{align*}
  We now appeal to the following lemma:
  \begin{lemma}
    \label{lem:err1}
    If the matrix $B \in \reals^{n \times n}$ is the adjacency matrix
    of a random Graph drawn from the Erd\"os-Renyi ensemble $G(n,d/n)$
    with $d \geq C \log n$ and $D$ is the corresponding diagonal
    matrix whose entry $d_{ii}$ is equal to the degree of node $i$,
    then we have that
    \begin{equation*}
      \|B/d - 1 1^T/n\|_2 \leq C {\sqrt{\frac{\log n}{d}}}
    \end{equation*}
    and
    \begin{equation*}
      \|B/d - D^{-1/2} B D^{-1/2}\|_2 \leq C {\sqrt{\frac{\log n}{d}}}
    \end{equation*}
    with probability at least $1 - 2 n^{-C n/(n-d)/8}$.
  \end{lemma}
  At this point, applying the
  above bound yields the result. It remains to prove the above bound.
\subsection{Proof of Lemma~\ref{lem:err1}}
\label{app:err1}
We prove the result in two parts. We first focus on establishing that
\begin{equation*}
  \| B/d - 1 1^T/n\|_2 \leq C {\sqrt{\frac{\log n}{d}}}
\end{equation*}
with probability at least $1 - n^{-Cn/(n-d)8}$. To prove this result, we
appeal to the following
\begin{lemma}[Theorem 1.4~\citep{Jo11}]
  Consider a finite sequence $\{X_k\}$ of independent, random, self-adjoint matrices
  with dimension $d$. Assume that each random matrix satisfies
  \begin{equation*}
    \E X_k = 0 \quad \text{ and } \quad \lambda_{\max}(X_k) \leq R \quad \text{almost surely}.
  \end{equation*}
  Then, for all $t \geq 0$,
  \begin{equation*}
    \prob \left ( \lambda_{\max}\left ( \sum_k X_k \right ) \geq t \right ) \; \leq
    d \cdot \exp \left ( \frac{-t^2/2}{\sigma^2 + R t/3}\right ) \quad \text{where $\sigma^2 = \|\sum_k \E X_k^2 \|_2$.}
  \end{equation*}
\end{lemma}
In our setting we are interested in the random matrix $B$ where we can write $B$ as
\begin{equation*}
  B - 1 1^T d/n= \sum_{i>j} (A_{ij} - d/n) (e_i e_j^T + e_j e_i^T) + \sum_i (A_{ii} - d/n) e_i e_i^T
\end{equation*}
where $A_{ij}$ is a Bernoulli random variable with parameter
$d/n$. Therefore, in applying the above Lemma we have that $R=1$
almost surely and $\sigma^2 = d(1-d/n)$. Setting $t = C \sqrt{d \log
  n}$ we have that
\begin{equation*}
  \| B/d - 1 1^T/n \|_2 \; \leq C \sqrt{\frac{\log n}{d}}
\end{equation*}
with probability at least $1 - n^{-Cn/(n-d)/8}$.

Next we show that
\begin{equation*}
  \|B/d - D^{-1/2} B D^{-1/2}\|_2 \leq C \sqrt{\frac{\log n}{d}}
\end{equation*}
with the same probability as above. To prove this result we will let
$E = D^{1/2} - d^{1/2} I$ and first note that
\begin{equation*}
  \|B/d - D^{-1/2} B D^{-1/2}\|_2 \leq \frac{1}{d \cdot \dmin} \|D^{1/2} B D^{1/2} - d B\|_2
\end{equation*}
because $\|D^{1/2}\|_2 = \frac{1}{\dmin}$. Some simple calculations show that
\begin{equation*}
  \|D^{1/2} B D^{1/2} - d B\|_2 \leq \|B\|_2 \cdot \left [ \|E\|_2^2 + 2 d^{1/2} \|E\|_2 \right ]
\end{equation*}
by above we know that $\|B\|_2 \leq 2d$ with high probability. Therefore,
\begin{equation*}
  \|B/d - D^{-1/2} B D^{-1/2}\|_2 \leq \frac{2}{\dmin} \left [ \|E\|_2^2 + 2 d^{1/2} \|E\|_2 \right ]
\end{equation*}
An application of Bernstein's inequality shows that with probability at least
$1 - 2 n^{-C n/(n-d)/8}$ we have $\|E \|_2 \leq 10 C \sqrt{\log n}$. Finally,
using the fact that with high probability $\dmin \geq \frac{1}{2} d$
\begin{equation*}
  \|B/d - D^{-1/2} B D^{-1/2}\|_2 \leq 12 C \sqrt{\frac{\log n}{d}}
\end{equation*}
with probability at least $1 - 2 n^{-C n/(n-d)/8}$.
\end{APPENDIX}
%
%




\bibliographystyle{ormsv080} 
\bibliography{ranking} 

\begin{thebibliography}{53}
\expandafter\ifx\csname natexlab\endcsname\relax\def\natexlab#1{#1}\fi
\expandafter\ifx\csname url\endcsname\relax
  \def\url#1{{\tt #1}}\fi
\expandafter\ifx\csname urlprefix\endcsname\relax\def\urlprefix{URL }\fi
\expandafter\ifx\csname urlstyle\endcsname\relax
  \expandafter\ifx\csname doi\endcsname\relax
  \def\doi#1{doi:\discretionary{}{}{}#1}\fi \else
  \expandafter\ifx\csname doi\endcsname\relax
  \def\doi{doi:\discretionary{}{}{}\begingroup \urlstyle{rm}\Url}\fi \fi

\bibitem[{Adler et~al.(1994)Adler, Gemmell, Harchol-Balter, Karp, and
  Kenyon}]{adleretal}
Adler, M., P.~Gemmell, M.~Harchol-Balter, R.~M. Karp, C.~Kenyon. 1994.
\newblock Selection in the presence of noise: the design of playoff systems.
\newblock {\it Proceedings of the fifth annual ACM-SIAM symposium on Discrete
  algorithms\/}. SODA '94, Society for Industrial and Applied Mathematics,
  564--572.

\bibitem[{Ailon(2010)}]{Ail10}
Ailon, N. 2010.
\newblock Aggregation of partial rankings, p-ratings and top-m lists.
\newblock {\it Algorithmica\/} {\bf 57}(2) 284--300.

\bibitem[{Ailon et~al.(2008)Ailon, Charikar, and Newman}]{ACN08}
Ailon, N., M.~Charikar, A.~Newman. 2008.
\newblock Aggregating inconsistent information: ranking and clustering.
\newblock {\it Journal of the ACM (JACM)\/} {\bf 55}(5) 23.

\bibitem[{Altman and Tennenholtz(2005)}]{AT05}
Altman, A., M.~Tennenholtz. 2005.
\newblock Ranking systems: the pagerank axioms.
\newblock {\it Proceedings of the 6th ACM conference on Electronic commerce\/}.
  ACM, 1--8.

\bibitem[{Ammar and Shah(2011)}]{AS11}
Ammar, A., D.~Shah. 2011.
\newblock Ranking: Compare, don't score.
\newblock {\it Communication, Control, and Computing (Allerton), 2011 49th
  Annual Allerton Conference on\/}. 776--783.

\bibitem[{Arrow(1963)}]{Arrow53}
Arrow, K.~J. 1963.
\newblock {\it Social Choice and Individual Values\/}.
\newblock Yale University Press.

\bibitem[{Boyd et~al.(2005)Boyd, Ghosh, Prabhakar, and Shah}]{BGPS05}
Boyd, S., A.~Ghosh, B.~Prabhakar, D.~Shah. 2005.
\newblock Mixing times for random walks on geometric random graphs.
\newblock {\it SIAM ANALCO\/} .

\bibitem[{Bradley and Terry(1955)}]{BT55}
Bradley, R.~A., M.~E. Terry. 1955.
\newblock Rank analysis of incomplete block designs: I. the method of paired
  comparisons.
\newblock {\it Biometrika\/} {\bf 39}(3/4) 324--345.

\bibitem[{Braverman and Mossel(2008)}]{BM08}
Braverman, M., E.~Mossel. 2008.
\newblock Noisy sorting without resampling.
\newblock {\it Proceedings of the nineteenth annual ACM-SIAM symposium on
  Discrete algorithms\/}. SODA '08, Society for Industrial and Applied
  Mathematics, 268--276.

\bibitem[{Brin and Page(1998)}]{BP98}
Brin, S., L.~Page. 1998.
\newblock The anatomy of a large-scale hypertextual web search engine.
\newblock {\it Seventh International World-Wide Web Conference (WWW 1998)\/}.

\bibitem[{Cand\`es and Recht(2009)}]{CR09}
Cand\`es, E.~J., B.~Recht. 2009.
\newblock Exact matrix completion via convex optimization.
\newblock {\it Foundations of Computational Mathematics\/} {\bf 9}(6) 717--772.

\bibitem[{Condorcet(1785)}]{Co1785}
Condorcet, M. 1785.
\newblock {\it Essai sur l'application de l'analyse \`a la probabilit\'e des
  d\'ecisions rendues \`a la pluralit\'e des voix\/}.
\newblock l'Imprimerie Royale.

\bibitem[{David(1963)}]{Dav63}
David, H.~A. 1963.
\newblock {\it The method of paired comparisons\/}, vol.~12.
\newblock DTIC Document.

\bibitem[{{de Borda}(1781)}]{Borda1781}
{de Borda}, J.~C. 1781.
\newblock M{\'e}moire sur les {\'e}lections au scrutin .

\bibitem[{Diaconis and Saloff-Coste(1993)}]{DS93}
Diaconis, P., L.~Saloff-Coste. 1993.
\newblock Comparison theorems for reversible markov chains.
\newblock {\it The Annals of Applied Probability\/} {\bf 3}(3) 696--730.

\bibitem[{Duchi et~al.(2010)Duchi, Mackey, and Jordan}]{DucMacJor10}
Duchi, J.~C., L.~Mackey, M.~I. Jordan. 2010.
\newblock On the consistency of ranking algorithms.
\newblock {\it Proceedings of the ICML Conference\/}. Haifa, Israel.

\bibitem[{Dwork et~al.(2001{\natexlab{a}})Dwork, Kumar, Naor, and
  Sivakumar}]{dwork01}
Dwork, C., R.~Kumar, M.~Naor, D.~Sivakumar. 2001{\natexlab{a}}.
\newblock Rank aggregation methods for the web.
\newblock {\it Proceedings of the Tenth International World Wide Web
  Conference, 2001\/}.

\bibitem[{Dwork et~al.(2001{\natexlab{b}})Dwork, Kumar, Naor, and
  Sivakumar}]{DKNS01}
Dwork, Cynthia, Ravi Kumar, Moni Naor, Dandapani Sivakumar. 2001{\natexlab{b}}.
\newblock Rank aggregation methods for the web.
\newblock {\it Proceedings of the 10th international conference on World Wide
  Web\/}. ACM, 613--622.

\bibitem[{Farnoud et~al.(2012)Farnoud, Touri, and Milenkovic}]{FTO12}
Farnoud, F., B.~Touri, O.~Milenkovic. 2012.
\newblock Novel distance measures for vote aggregation.
\newblock {\it arXiv preprint arXiv:1203.6371\/} .

\bibitem[{Gleich and Lim(2011)}]{GL11}
Gleich, D.~F., L.~Lim. 2011.
\newblock Rank aggregation via nuclear norm minimization.
\newblock {\it Proceedings of the 17th ACM SIGKDD international conference on
  Knowledge discovery and data mining\/}. ACM, 60--68.

\bibitem[{Guiver and Snelson(2009)}]{GS09}
Guiver, J., E.~Snelson. 2009.
\newblock Bayesian inference for plackett-luce ranking models.
\newblock {\it Proceedings of the 26th Annual International Conference on
  Machine Learning\/}. ACM, 377--384.

\bibitem[{Hajek et~al.(2014)Hajek, Oh, and Xu}]{HOX14}
Hajek, Bruce, Sewoong Oh, Jiaming Xu. 2014.
\newblock Minimax-optimal inference from partial rankings.
\newblock {\it Advances in neural information processing systems (NIPS)\/}.

\bibitem[{Hochbaum(2006)}]{Hoch06}
Hochbaum, D.~S. 2006.
\newblock Ranking sports teams and the inverse equal paths problem.
\newblock {\it Internet and Network Economics\/}. Springer, 307--318.

\bibitem[{Horn and Johnson(1985)}]{HJ85}
Horn, R.~A., C.~R. Johnson. 1985.
\newblock {\it Matrix Analysis\/}.
\newblock Cambridge University Press.

\bibitem[{Hunter(2004)}]{Hun04}
Hunter, David~R. 2004.
\newblock Mm algorithms for generalized bradley-terry models.
\newblock {\it Annals of Statistics\/}  384--406.

\bibitem[{Jiang et~al.(2011)Jiang, Lim, Yao, and Ye}]{JLYY11}
Jiang, X., L.~Lim, Y.~Yao, Y.~Ye. 2011.
\newblock Statistical ranking and combinatorial hodge theory.
\newblock {\it Mathematical Programming\/} {\bf 127}(1) 203--244.

\bibitem[{Kamvar et~al.(2003)Kamvar, Schlosser, and Garcia-Molina}]{KSG03}
Kamvar, S.~D., M.~T. Schlosser, H.~Garcia-Molina. 2003.
\newblock The eigentrust algorithm for reputation management in p2p networks.
\newblock {\it Proceedings of the 12th international conference on World Wide
  Web\/}. WWW '03, ACM, New York, NY, USA, 640--651.

\bibitem[{Keener(1993)}]{Kee93}
Keener, J.~P. 1993.
\newblock The perron-frobenius theorem and the ranking of football teams.
\newblock {\it SIAM review\/} {\bf 35}(1) 80--93.

\bibitem[{Kendall(1955)}]{Ken55}
Kendall, M.~G. 1955.
\newblock Further contributions to the theory of paired comparisons.
\newblock {\it Biometrics\/} {\bf 11}(1) 43--62.

\bibitem[{Kendall and Smith(1940)}]{KS40}
Kendall, M.~G., B.~B. Smith. 1940.
\newblock On the method of paired comparisons.
\newblock {\it Biometrika\/}  324--345.

\bibitem[{Keshavan et~al.(2010)Keshavan, Montanari, and Oh}]{KMO10}
Keshavan, R.~H., A.~Montanari, S.~Oh. 2010.
\newblock Matrix completion from noisy entries.
\newblock {\it Journal of Machine Learning Research\/} {\bf 11} 2057--2078.

\bibitem[{L.~R.~Ford(1957)}]{Ford57}
L.~R.~Ford, Jr. 1957.
\newblock Solution of a ranking problem from binary comparisons.
\newblock {\it The American Mathematical Monthly\/} {\bf 64}(8) 28--33.

\bibitem[{Lu and Boutilier(2011)}]{LB11}
Lu, T., C.~Boutilier. 2011.
\newblock Learning mallows models with pairwise preferences.
\newblock {\it Proceedings of the 28th International Conference on Machine
  Learning (ICML-11)\/}. 145--152.

\bibitem[{Luce(1959)}]{Luce59}
Luce, D.~R. 1959.
\newblock {\it {Individual Choice Behavior}\/}.
\newblock Wiley, New York.

\bibitem[{Mallows(1957)}]{Mallows57}
Mallows, C.~L. 1957.
\newblock Non-null ranking models. i.
\newblock {\it Biometrika\/}  114--130.

\bibitem[{McFadden(1973)}]{McF73}
McFadden, D. 1973.
\newblock Conditional logit analysis of qualitative choice behavior.
\newblock {\it Frontiers in Econometrics\/}  105--142.

\bibitem[{Mosteller(1951)}]{Mos51}
Mosteller, F. 1951.
\newblock Remarks on the method of paired comparisons: I. the least squares
  solution assuming equal standard deviations and equal correlations.
\newblock {\it Psychometrika\/} {\bf 16}(1) 3--9.

\bibitem[{Negahban and Wainwright(2012)}]{NW11}
Negahban, S., M.~J. Wainwright. 2012.
\newblock Restricted strong convexity and (weighted) matrix completion: Optimal
  bounds with noise.
\newblock {\it Journal of Machine Learning Research\/}  1665--1697.

\bibitem[{Newman(2010)}]{newman}
Newman, M.~E.~J. 2010.
\newblock {\it Networks: An Introduction\/}.
\newblock Oxford University Press.

\bibitem[{Osting et~al.(2013)Osting, Brune, and Osher}]{OBO13}
Osting, B., C.~Brune, S.~Osher. 2013.
\newblock Enhanced statistical rankings via targeted data collection.
\newblock {\it Proceedings of the 30th International Conference on Machine
  Learning\/}. 489--497.

\bibitem[{Plackett(1975)}]{Pla75}
Plackett, R.~L. 1975.
\newblock The analysis of permutations.
\newblock {\it Applied Statistics\/}  193--202.

\bibitem[{Rajkumar and Agarwal(2014)}]{RA14}
Rajkumar, A., S.~Agarwal. 2014.
\newblock A statistical convergence perspective of algorithms for rank
  aggregation from pairwise data.
\newblock {\it Proceedings of The 31st International Conference on Machine
  Learning\/}. 118--126.

\bibitem[{Rao(1945)}]{Rao45}
Rao, C.~R. 1945.
\newblock Information and accuracy attainable in the estimation of statistical
  parameters.
\newblock {\it Bulletin of the Calcutta Mathematical Society\/} {\bf 37}(3)
  81--91.

\bibitem[{Saaty(2003)}]{Saa03}
Saaty, T.~L. 2003.
\newblock Decision-making with the ahp: Why is the principal eigenvector
  necessary.
\newblock {\it European Journal of Operational Research\/} {\bf 145} pp.
  85--91.

\bibitem[{Salganik and Levy(2012)}]{SL12}
Salganik, M.~J., K.~E.C. Levy. 2012.
\newblock Wiki surveys: Open and quantifiable social data collection.
\newblock Tech. Rep.~{\tt arXiv:1202.0500}.

\bibitem[{Seeley(1949)}]{See49}
Seeley, J.~R. 1949.
\newblock The net of reciprocal influence.
\newblock {\it Canadian Journal of Psychology\/} {\bf 3}(4) 234--240.

\bibitem[{Shah and Zaman(2011)}]{SZ11}
Shah, D., T.~Zaman. 2011.
\newblock Rumors in a network: who?s the culprit?
\newblock {\it IEEE Transactions on Information Theory\/} {\bf 57}(8)
  5163--5181.

\bibitem[{Shah and Zaman(2015)}]{SZ15}
Shah, D., T.~Zaman. 2015.
\newblock Finding rumor sources on random trees.
\newblock {\it Operations Research\/} .

\bibitem[{Talluri and VanRyzin(2005)}]{VanRyzin}
Talluri, K.~T., G.~VanRyzin. 2005.
\newblock {\it The Theory and Practice of Revenue Management\/}.
\newblock springer.

\bibitem[{Tropp(2011)}]{Jo11}
Tropp, J. 2011.
\newblock User-friendly tail bounds for sums of random matrices.
\newblock {\it Foundations of Computational Mathematics\/} .

\bibitem[{Vigna(2009)}]{Vig09}
Vigna, S. 2009.
\newblock Spectral ranking.
\newblock {\it arXiv preprint arXiv:0912.0238\/} .

\bibitem[{Volkovs and Zemel(2012)}]{VZ12}
Volkovs, M.~N., R.~S. Zemel. 2012.
\newblock A flexible generative model for preference aggregation.
\newblock {\it Proceedings of the 21st international conference on World Wide
  Web\/}. ACM, 479--488.

\bibitem[{Wei(1952)}]{Wei52}
Wei, T.~H. 1952.
\newblock The algebraic foundations of ranking theory.
\newblock Ph.D. thesis, University of Cambridge.

\end{thebibliography}





\end{document}